\documentclass[10pt,journal]{IEEEtran}
\ifCLASSOPTIONcompsoc
  \usepackage[nocompress]{cite}
\else
  \usepackage{cite}
\fi

\usepackage{algorithm}
\usepackage{algorithmic}
\usepackage{soul}
\usepackage{epsfig}
\usepackage{graphicx}
\usepackage{amsmath}
\usepackage{amssymb}

\newcommand{\etal}{\textit{et al}.}
\newcommand{\ie}{\textit{i}.\textit{e}.~}
\newcommand{\eg}{\textit{e}.\textit{g}.~}
\newcommand{\vs}{\textit{v}.\textit{s}.~}

\usepackage{ifpdf}
\usepackage{graphicx} %%Grafiken und normales LaTeX
\usepackage{diagbox}
\usepackage{multirow}
\usepackage{color}
\usepackage{subfigure}
\usepackage{enumitem}
\usepackage{setspace}
\usepackage{booktabs}
\usepackage{url}
\usepackage{color, xcolor}
\definecolor{myblue}{rgb}{0.6,0.7,0.95}

\ifCLASSINFOpdf
\else
\fi

\begin{document}
\title{Weakly Aligned Feature Fusion for Multi-modal Object Detection}

\author{Lu~Zhang,~Zhiyong~Liu,~\IEEEmembership{Senior Member,~IEEE,}~Xiangyu~Zhu,~Zhan~Song,~Xu~Yang,\\~Zhen~Lei,~\IEEEmembership{Senior Member,~IEEE,}~and~Hong~Qiao,~\IEEEmembership{Fellow,~IEEE}\\

\thanks{This work was supported by the National Key Research and Development Plan of China under Grant 2020AAA0108902, the Strategic Priority Research Program of Chinese Academy of Science under Grant XDB32050100, the Beijing Science and Technology Plan Project  under grant Z201100008320029, the NSFC under Grants 61627808, 61876178, and 61806196, the Dongguan Core Technology Research Frontier
Project, China (2019622101001).}
\thanks{L. Zhang, Z. Liu, X. Yang and H. Qiao are with the State Key Laboratory of Management and Control for Complex Systems, Institute of Automation, Chinese Academy of Sciences, Beijing 100190, China, and also with University of Chinese Academy of Sciences, Beijing 100086, China. Z. Liu and X. Yang are also with Center for Excellence in Brain Science and Intelligence Technology, Chinese Academy of Sciences, Shanghai 200031, China. E-mail: {\{zhanglu2016, zhiyong.liu, xu.yang, hong.qiao\}@ia.ac.cn}}
\thanks{X. Zhu and Z. Lei are with the Center for Biometrics and Security Research \& National Laboratory of Pattern Recognition, Institute of Automation, Chinese Academy of Sciences, Beijing 100190, China, and also with University of Chinese Academy of Sciences, Beijing 100086, China. Z. Lei is also with the Centre for Artificial Intelligence and Robotics, Hong Kong Institute of Science \& Innovation, Chinese Academy of Sciences, Hong Kong, China. Email: \{xiangyu.zhu, zlei\}@nlpr.ia.ac.cn}
\thanks{Z. Song is with the Shenzhen Institutes of Advanced Technology, Chinese Academy of Sciences, Shenzhen 518055, Guangdong, China. E-mail: zhan.song@siat.ac.cn}
}% <-this % stops an unwanted space

\maketitle

\begin{abstract}
To achieve accurate and robust object detection in the real-world scenario, various forms of images are incorporated, such as color, thermal, depth, etc. However, multi-modal data often suffer from the position shift problem, i.e. the image pair is not strictly aligned, making one object has different positions in different modalities. For the deep learning method, this problem makes it difficult to fuse multi-modal features and puzzles the CNN training. In this paper, we propose a general multi-modal detector named Aligned Region CNN (AR-CNN) to tackle the position shift problem. Firstly, a region feature alignment module with adjacent similarity constraint is designed to consistently predict the position shift between two modalities and adaptively align the cross-modal region features. Secondly, we propose a novel RoI jitter strategy to improve the robustness to unexpected shift patterns. Thirdly, we present a new multi-modal feature fusion method, which selects the more reliable feature and suppresses the less useful one via feature re-weighting. Additionally, by locating bounding boxes in both modalities and building their relationships, we provide novel multi-modal labelling named KAIST-Paired. Extensive experiments on 2D and 3D object detection, RGB-T and RGB-D datasets demonstrate the effectiveness and robustness of our method. 
\end{abstract}


\IEEEdisplaynontitleabstractindextext

% \fi
\IEEEpeerreviewmaketitle

\section{Introduction}\label{sec:introduction}


\begin{figure}
\centering
\includegraphics[width=3.2in]{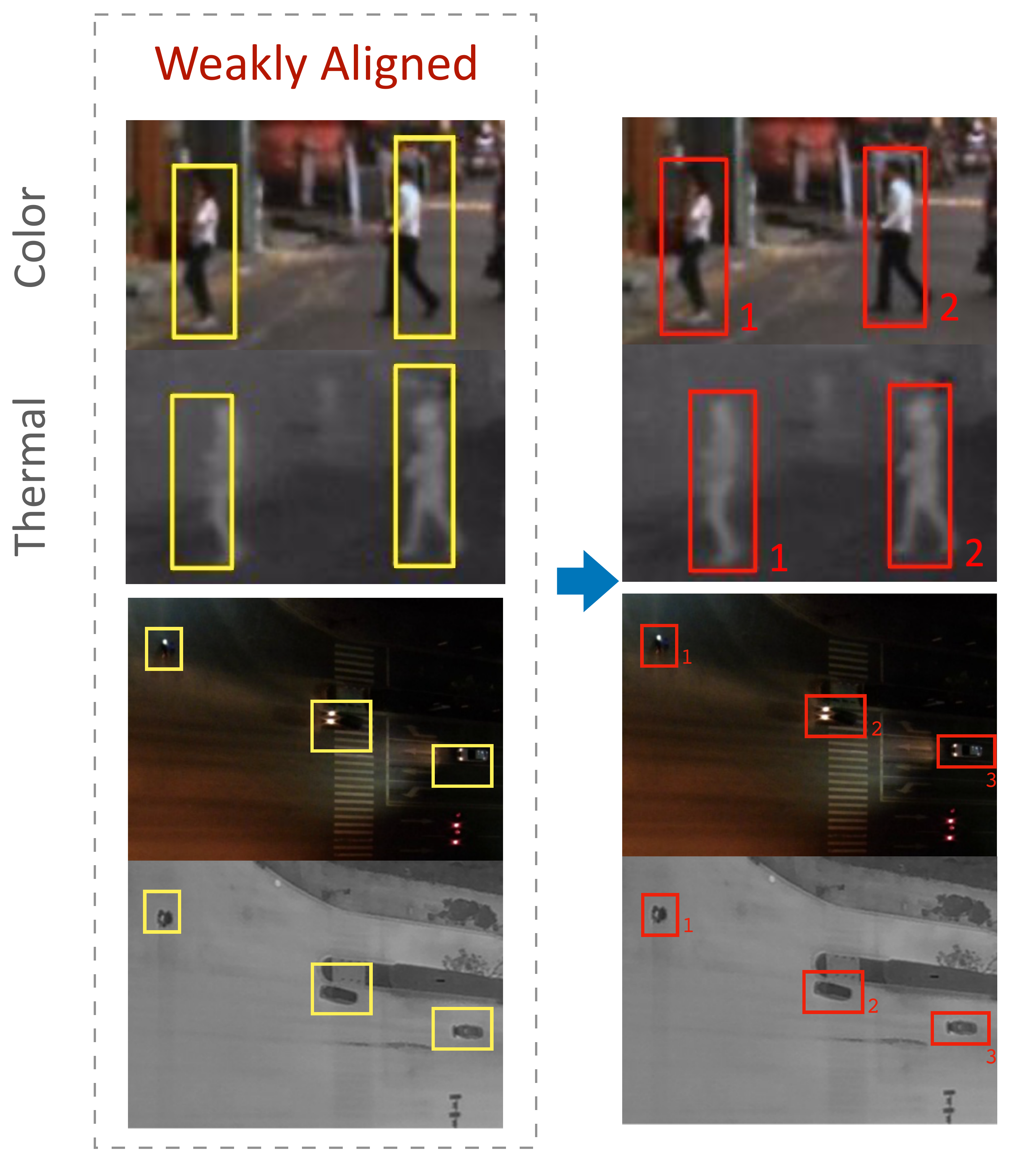}   %1.84,1.21

\caption{Illustration of the weakly aligned image pair. Yellow boxes in the left column denote original annotations shared by two modalities, which present the position shift in multi-modal inputs. Red boxes in the right column denote the proposed multi-modal labelling, which has pairs of annotations that match each input modality.}% 
\label{figure-weakly-aligned}                     
\end{figure}

\IEEEPARstart{R}{obust} object detection is a crucial ingredient of many computer vision applications in the real world, such as robotics, surveillance, and autonomous driving. 
Nowadays, smart devices are usually equipped with different sensors (e.g. cameras, thermal cameras, time-of-flight, structured light, LiDAR, Radars), which provide different image forms. Compared to pure RGB data \cite{lin2014microsoft, xia2018dota, pang2020tju}, integrating the extra modality is proven to be effective for recognition: infrared camera provides biometric information for around-the-clock pedestrian detection \cite{hwang2015multispectral, gonzalez2016pedestrian, yuan2015multi} and face recognition \cite{klare2012heterogeneous, shao2016cross}, and depth channel provides 3D information for object detection \cite{gupta2014learning, eitel2015multimodal, deng2017amodal} and pose estimation \cite{schwarz2015rgb, choi20123d}, etc. Motivated by this, the multi-modal object detection has attracted massive attention, not only towards the improvement of accuracy but also for a better robustness promise.

Most existing multi-modal methods \cite{feng2019deep, hwang2015multispectral, deng2017amodal} detect objects under the \textit{alignment assumption}, i.e. the image pairs from different modalities are well aligned and have strong pixel-to-pixel correspondence. However, this assumption does not always hold in practice, as shown in Figure \ref{figure-weakly-aligned}, which poses great challenges to existing methods. To fill the gap with the alignment assumption and real-world applications, we discuss three main reasons behind the challenges.

First, the alignment quality of image pair is limited by the physical properties of different sensors, e.g. geometrical disparity due to relative displacement \cite{barnard1980disparity}, mismatched resolutions and field-of-views. For example, the border areas of color images need to be sacrificed since the color camera usually has a higher resolution and larger field view than the thermal camera \cite{hwang2015multispectral}. This will naturally lead to imperfect calibration and result in alignment error. 
Second, the calibration and re-calibration processes are important while tortuous, generally require particular hardware such as a special heated calibration board \cite{kim2015geometrical, treible2017cats, choi2018kaist}, making re-calibration difficult when the device is in operation. 
Moreover, the alignment is easily impacted by external disturbance (e.g. mechanical vibrations and temperature variation) and hardware aging, which is hard to be avoided when a system starts to operate. 

As a result, the multi-modal images are often weakly aligned in practice, thus there is position shift between one object in different modalities, i.e. the \textit{position shift} problem. 
The position shift problem degrades the CNN-based detectors mainly in two aspects.  
First, multi-modal features in the corresponding position are spatially mismatched, which will puzzle the CNN training. 
Second, the position shift problem makes it difficult to match objects from both input modalities by using a shared bounding box. As a workaround, existing multi-modal datasets \cite{hwang2015multispectral, gonzalez2016pedestrian} use larger bounding boxes to encompass objects from both modalities or annotate objects on a single input modality. 
Such label bias can lead to bad supervision signals and is harmful to the training process, especially for modern detectors \cite{ren2015faster, liu2016ssd, zhao2019object, li2021improving} since they generally use the intersection over union (IoU) overlap for foreground/background classification. 

Thus, how to robustly locate the object on weakly aligned modalities remains to be a crucial issue for multi-modal object detection, while it is barely touched or studied in the literature. Besides, due to different devices, deployment environments, and operation duration, the priorities of modality information and the patterns of position shifts are ever-changing. Hence it is a very important challenge to cope with the position shift problem using a single model. 

In this paper, a novel framework is proposed to tackle the position shift problem in an end-to-end fashion. Specifically, we present a novel Region Feature Alignment (RFA) module to shift and align the to-be-fused region features from input modalities. With this module, the region-based alignment process is inserted in the network and works in a learnable way. To further enhance the robustness to different patterns of position shift, we propose the RoI jitter training strategy, which augments the RoIs of the sensed modality via random jitter.

To validate the effectiveness of our method, we apply the proposed framework on the typical \textit{RGB-T (multispectral) pedestrian detection}. To better meet the around-the-clock demands, we design a new confidence-aware fusion module, which selects the more reliable feature and suppresses the less useful one via adaptive feature re-weighting. Besides, we provide a novel KAIST-Paired\footnote{available at \url{https://github.com/luzhang16/AR-CNN}} annotation by locating the bounding boxes in both modalities and building their relationships. We also collect a drone-based RGB-T dataset, which includes more object categories (\eg car, bus, truck, cyclist), to validate the generalization to RGB-T object detection. Moreover, we extend the proposed method to the \textit{RGB-D based object detection} task. Experiments on 3D object detection are conducted on the standard NYUv2 dataset. To further validate the proposed framework, we build a dataset containing different kinds of indoor objects, called SL-RGBD, in which image pairs are labelled with multi-modal bounding boxes. Extensive experimental results demonstrate that the proposed method significantly improves the robustness to position shift problem and takes full advantage of multi-modal inputs by effective feature fusion, thus achieving state-of-the-art result on the challenging KAIST and CVC-14 datasets. For the general 2D and 3D object detection tasks, our method also achieves the best robustness performance on the NYUv2 and SL-RGBD datasets. 

This paper is an extension of our previous work \cite{Zhang_2019_ICCV} mainly in the following aspects: 
1) We improve our method with the adjacent similarity constraint for the region shift prediction and extend the approach to a general multi-modal object detection task, including RGB-Thermal and RGB-Depth detection, 2D and 3D object detection; 2) The experimental upper bound of position shift for the weakly aligned image pair are introduced to specify the definition of the \textit{weakly} aligned;  3) Additional experiments and analyses are conducted to better show the effectiveness and generalization ability of the proposed method. 

The remainder of the paper is organized as follows: we review the related work in Section \ref{S2}. Section \ref{S3} discusses the position shift problem in weakly aligned image pairs and presents our motivation. Section \ref{S4} describes the proposed framework in detail. Extensive experiment results of RGB-T pedestrian detection are given in Section \ref{S5}. In Section \ref{S6}, we further report the results of the proposed method on RGB-D based 2D and 3D object detection. We conclude the paper in Section \ref{S7}.

\section{Related Work}
\label{S2}

Recent intelligence systems have access to various forms of images, such as RGB, depth, infrared, etc. Compared to the RGB-based method, the adoption and fusion of novel modality greatly improve the performance and robustness, thus enabling many practical applications, e.g. around-the-clock surveillance, autonomous driving and robot grasping.

\subsection{Multispectral (RGB-T) Pedestrian Detection} 
Pedestrian detection is an essential step for many applications and recently gets increasing attention. In previous years, many algorithms and features have been proposed, including the traditional detectors \cite{dollar2009integral, dollar2014fast, nam2014local, zhang2015filtered} and the lately dominated CNN-based detectors \cite{mao2017can, hosang2015taking, wang2018repulsion, zhang2018occlusion, xie2021psc}. With the recent release of large-scale multispectral pedestrian benchmarks, efficiently exploiting the multispectral data has shown great advantages on pedestrian detection, especially for the around-the-clock operation \cite{kim2018multispectral, choi2016thermal, choi2015all, MBNet-ECCV2020, wolpert2020anchor, zhang2020multispectral}. In \cite{hwang2015multispectral}, Hwang et al. propose the KAIST multispectral dataset and extend the ACF method to make full use of the aligned color-thermal image pairs.  With the success of deep learning, several CNN-based methods \cite{ wagner2016multispectral, zhang2019cross, guan2018exploiting, li2019illumination} are proposed in recent years. Liu et al.\cite{liu2016multispectral} build a two-stream detector and experimentally analyse different fusion timings. K{\"o}nig et al.\cite{konig2017fully} fuse features of the Region Proposal Network (RPN) and introduce the Boosted Forest (BF) framework as the classifier. In \cite{xu2017learning}, Xu et al. propose a cross-modality learning framework that can cope with missing thermal data at testing time. Zhang et al. \cite{zhang2019cross} utilize the cross-modality attention mechanism to recalibrate the channel responses of halfway feature maps.
However, most previous approaches conduct multi-modal fusion under the full alignment assumption. This not only hinders the use of weakly aligned datasets (such as CVC-14 \cite{gonzalez2016pedestrian}), but also limits the further development of multispectral pedestrian detection, which is worthy of attention but still lacks research.

\subsection{RGB-D Object Detection} 

We briefly introduce the 2.5D approaches in RGB-D images, in which the depth images are generally treated in a similar fashion as RGB images.  In \cite{spinello2011people}, the detector combines the Histogram of Oriented Gradients (HOG) features of RGB data and Histogram of Oriented Depths (HOD) of dense depth data in a probabilistic way. With the recent dominance of CNNs, Gupta et al. \cite{gupta2014learning} build a CNN-based detector on the top of pre-computed object segmentation candidates. They extend the R-CNN \cite{girshick2014rich} to utilize depth information with HHA encoding (Horizontal Disparity, Height above ground, and Angle of local surface normal w.r.t gravity direction). But the outputs of detectors are still limited to 2D ones. In order to infer the 3D bounding box, Deng et al. \cite{deng2017amodal} propose the Amodal3Det, which further explores the strong CNN to infer 3D dimensions in RGB-D data efficiently. Besides, the single-shot 3D-SSD \cite{luo20173d} is designed to achieve high-speed inference. \cite{rahman20193d} downsamples the RoIs to get better 3D initialization. 
Different from them, we consider the position shift problem in RGB-D pairs for 3D object detection.

\subsection{Weakly Aligned Image Pair}

Though the multi-modal object detection is extensively studied, few detection methods touch the position shift problem in a weakly aligned image pair. Since this problem is unavoidable in the realistic scenarios due to both hardware and environmental factors, before the detection methods, it can be independently mitigated in two ways: pre-process and post-process. 

The pre-process aims to solve the camera calibration problem and elaborate a delicate modality-aligned system. However, this hardware promise is easily impacted by unavoidable external disturbance. Also, the recalibration \cite{vidas2012mask, kim2015geometrical} process can be difficult due to the characteristics of additional modality. For example, the calibration of RGB and thermal images generally need particular hardware and the special heated calibration board \cite{kim2015geometrical , treible2017cats, choi2018kaist}.

A common paradigm of the post-process is to conduct image registration (i.e. spatial alignment) \cite{zitova2003image, brown1992survey, dawn2010remote}. As a image-level solution for the position shift problem, it geometrically aligns the \textit{reference} and \textit{sensed} images. This task mainly consists of four processes: feature detection, mapping function design, feature matching, image transformation and re-sampling. Although the image registration is well established, the low-level transformation on the whole image is often time-consuming and restricts end-to-end training of the CNN-based detector. Moreover, since different modalities present different appearances, the registration process faces great challenges because the correspondence of key points is harder to determine.

%------------------------------------------------------------------------
\section{Motivation}
\label{S3} 
In this section, we present our analysis of the position shift problem in the weakly aligned image pair. To provide insights into this problem, we firstly analyse the popular KAIST \cite{hwang2015multispectral} and CVC-14 \cite{gonzalez2016pedestrian} RGB-T pedestrian datasets, and the structured-light-based RGB-D system. Then we experimentally study how the position shift deteriorates the detection performance.  

\subsection{The Position Shift Problem}
\label{S3.1}
The position shift is defined as the spatial displacement between two images of different cameras. The image pair is taken in the same scene and time, generally including the same objects.
However, caused by the position shift problem, the pixel-to-pixel relationship does not hold, two corresponding image patches can be located in different positions.

\textbf{RGB-T Pair} From the image pairs in KAIST and CVC-14 RGB-T pedestrian datasets and the drone-based object dataset, we can observe several issues.

\begin{figure}
\subfigure[]{
\begin{minipage}[t]{0.53\linewidth}
\centering
\includegraphics[width=1.9in]{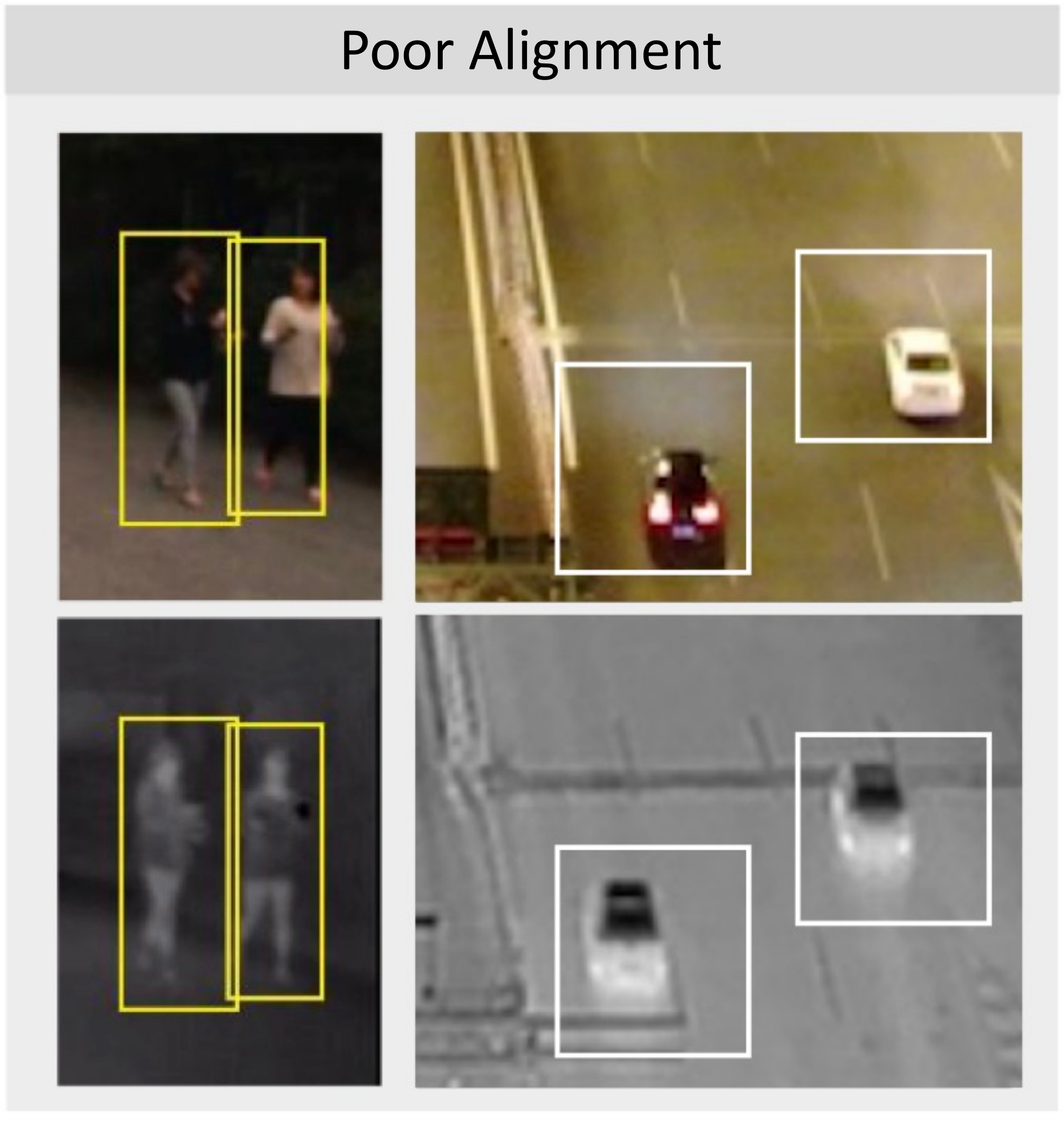}   
%\caption{fig1}
%\label{fig:side:a}
\end{minipage}%
\label{fig:side:a}}%
\subfigure[]{
\begin{minipage}[t]{0.45\linewidth}
\centering
\includegraphics[width=1.544in]{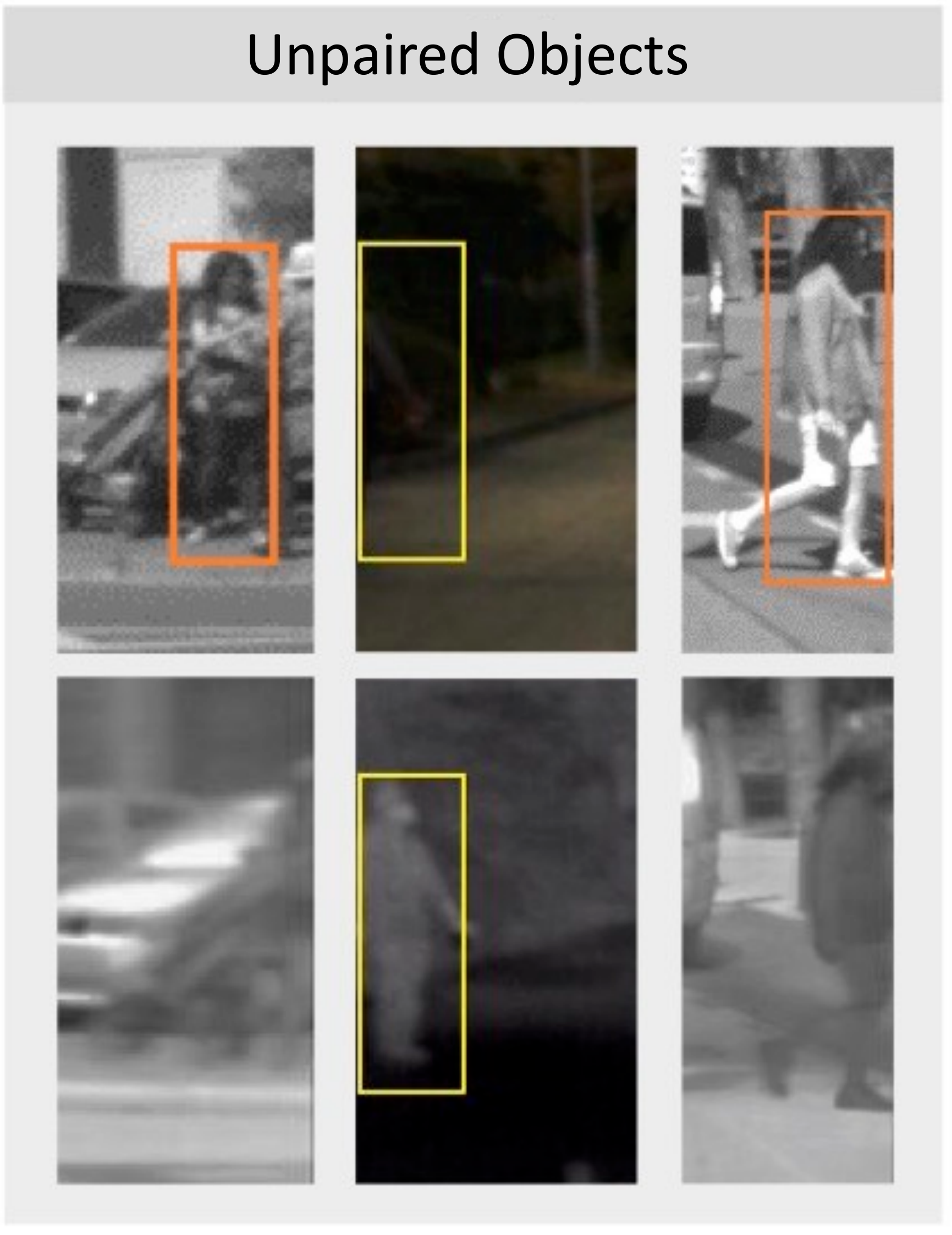} %1.544 w/o highlighted
%\caption{fig2}
%\label{fig:side:b}
\end{minipage}%
\label{fig:side:b}}
\caption{Some instances of original annotations in the KAIST (yellow boxes), CVC-14 (orange box), and drone-based (white boxes) datasets. We crop the image patches in the same position, but the objects pose poor alignment. Thus such a shared bounding box for both modalities can not precisely locate the object.}%
\label{figure-data-instance}                     
\end{figure}

\begin{figure}
\subfigure[KAIST dataset]{
\begin{minipage}[t]{0.49\linewidth}
\centering
\includegraphics[width=1.52in]{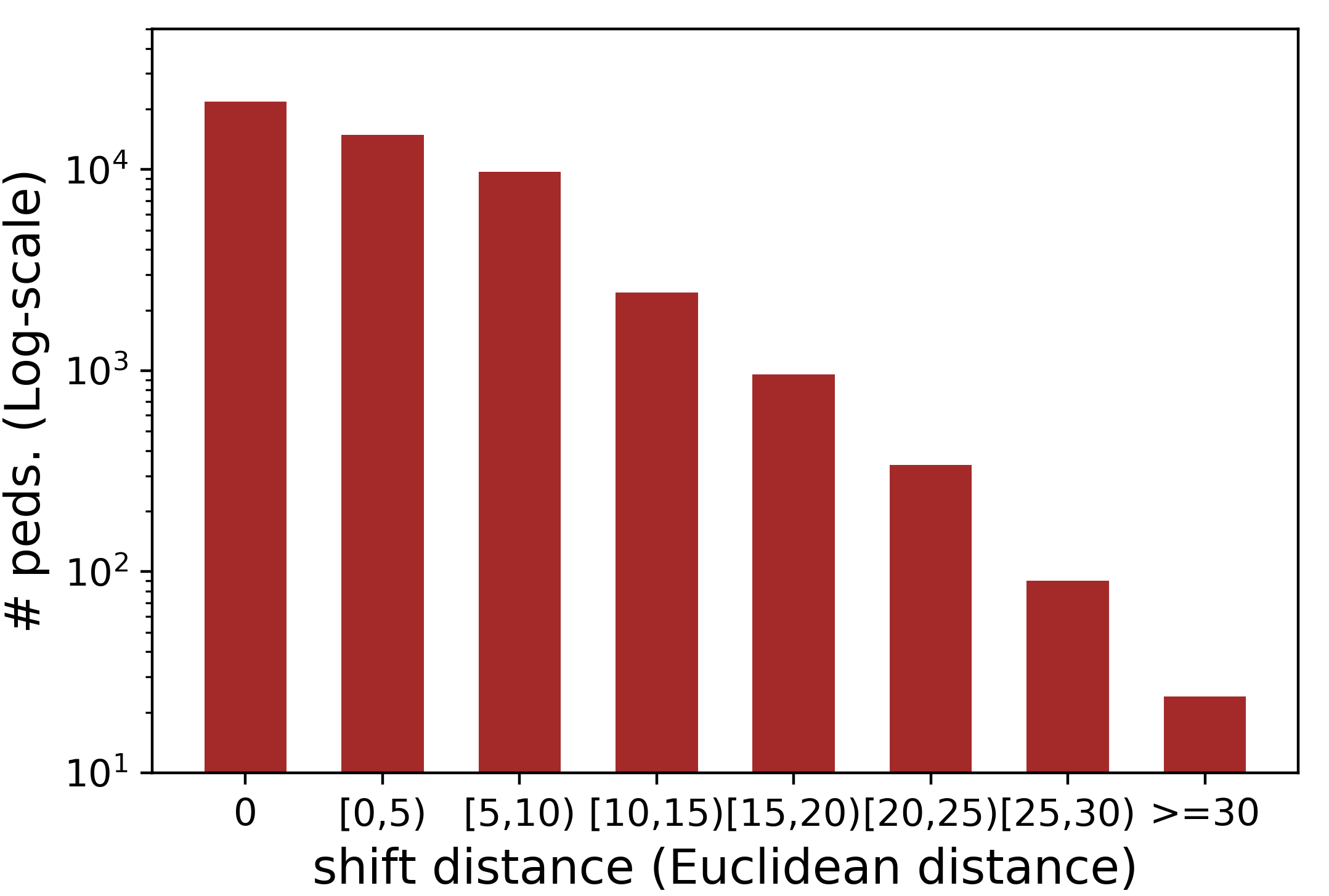}
%\caption{fig1}
\label{figure-stat-shifting-1}
\end{minipage}}%
\subfigure[CVC-14 dataset]{
\begin{minipage}[t]{0.51\linewidth}
\centering
\includegraphics[width=1.56in]{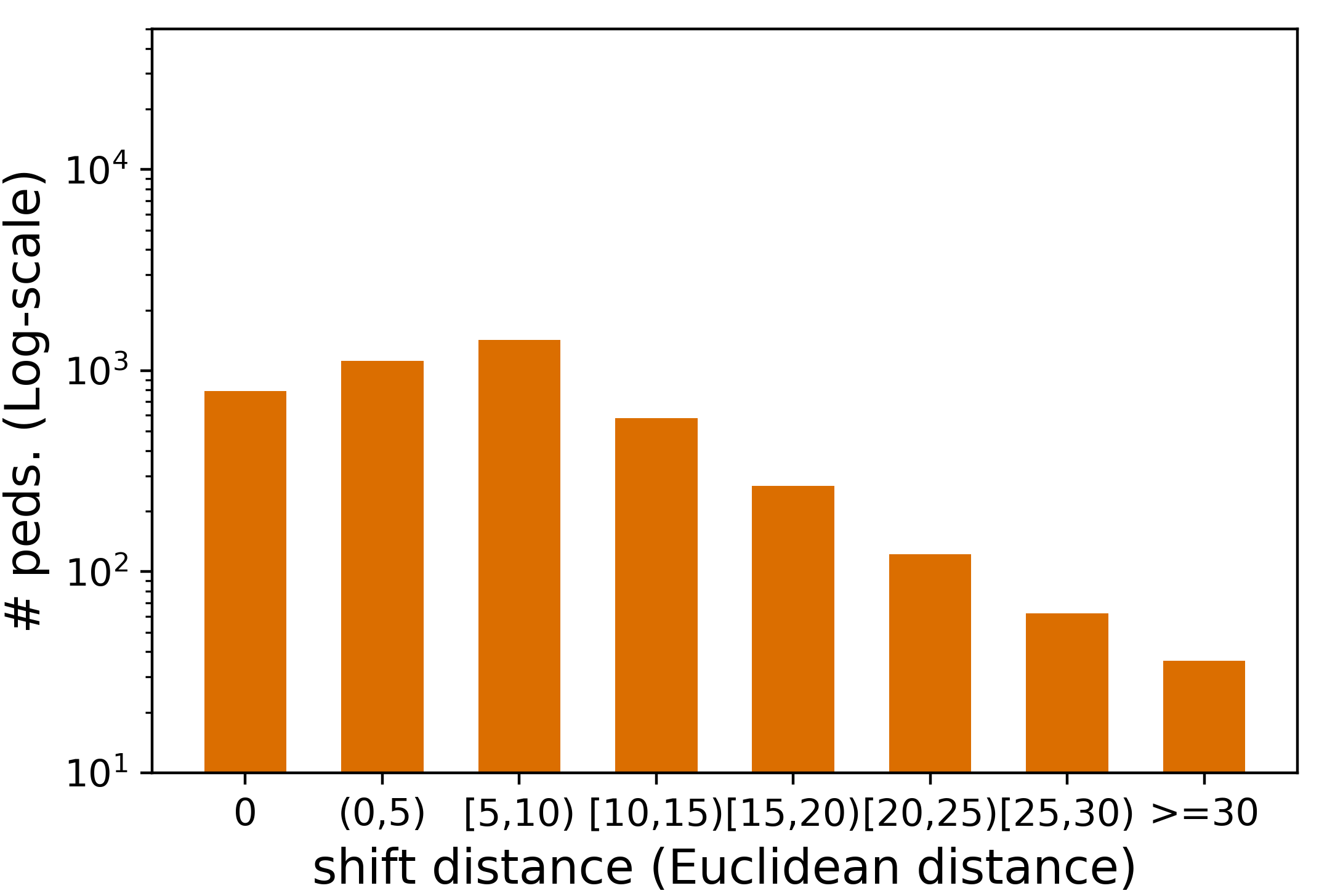}
%\caption{fig2}
\label{figure-stat-shifting-2}
\end{minipage}}%
\caption{Statistics of the position shift for bounding boxes in KAIST and CVC-14 datasets.} 
\label{figure-stat-shifting}                     
\end{figure}

\begin{itemize}
\item{\textit{Misaligned Features.}} 
Figure \ref{fig:side:a} illustrates the position shift between modalities, this problem makes it difficult to directly fuse such misaligned features. 
Besides, the shift varies in different positions even in a single image. In Figure \ref{figure-stat-shifting}, we present the statistics of position shift in two datasets, which show that collection devices and operation environments also influence the shift pattern.

\item{\textit{Localization Bias.}} 
As shown in Figure \ref{fig:side:a}, the annotations need to be adjusted to match the weakly aligned image pair.
One way to remedy is using larger bounding boxes to encompass objects from both modalities, but also including unnecessary background information. Another workaround is only focusing on one particular modality, but this could introduce bias for another modality.

\item{\textit{Unpaired Objects.}} 
In practice, color and thermal cameras often have different field-of-views, the situation is even worse when the synchronization and calibration are not good. 
Therefore, as shown in Figure \ref{fig:side:b}, the objects appear in one modality but are truncated or lost in another. 
Specifically, $\sim12.5\%$ ($2,245$ of $18,017$) of pedestrians are unpaired in CVC-14 \cite{gonzalez2016pedestrian}.
\end{itemize}

\textbf{RGB-D Pair} 
Around different types of 3D sensing systems, the projector-camera-based structured light methods have been more and more important for 3D vision related applications. A typical structured light system is composed of one projector and one industry camera. Since patterns used by the structured light systems are usually monochrome or binary, the monochrome cameras are preferred. 
In real applications, to obtain high-quality color texture for the reconstructed 3D models, an extra high-resolution RGB camera like a DSLR (Digital Single-Lens Reflex) camera is usually added to the system \cite{song2012accurate}. As a result, alignment of the high-resolution color image and the low-resolution point cloud or depth image is essential for any structured light systems which used extra color texture camera.

\begin{figure}

\includegraphics[width=3.4in]{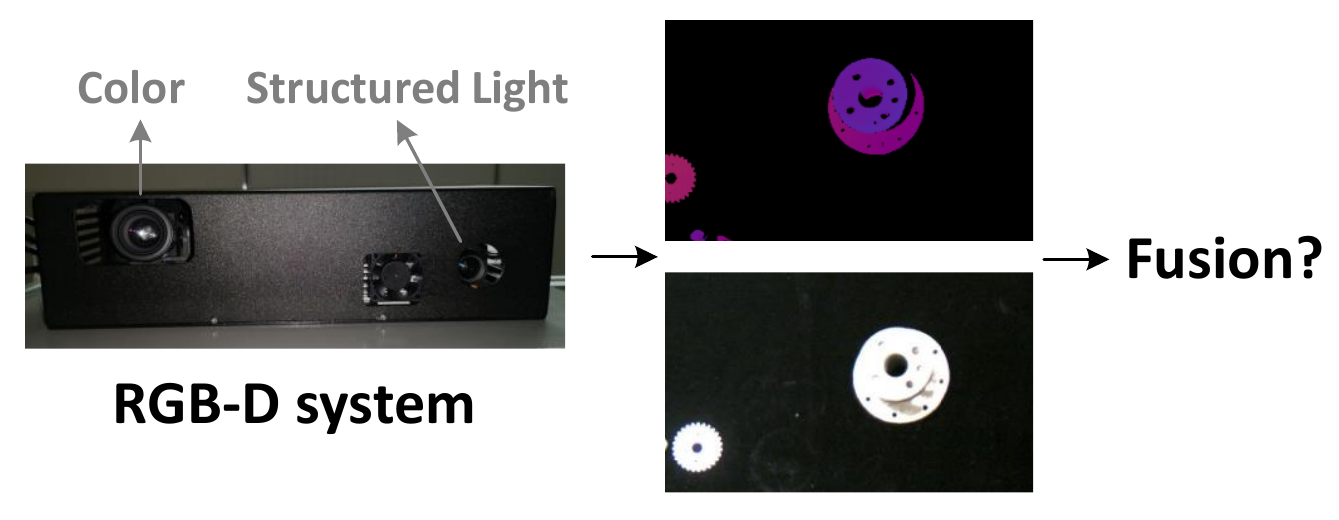}   %1.84,1.21

\caption{Illustration of the RGB-D system. The system consists of the color camera and the projector-camera-based structured light. As the calibration quality degrades, the RGB and depth image pair suffers from the position shift problem.}%
\label{figure-rgbd-sys}                     
\end{figure}

To reveal the realistic position shift problem of RGB-D data, we build such an RGB-D data collection system with a pair of color and structural light cameras, as shown in Figure \ref{figure-rgbd-sys}. We can see that even the system is well calibrated, time and external disturbance will degrade the quality of the calibration. As demonstrated in  Figure \ref{figure-rgbd-sys}, the main problem is position shift, along with milder deformation and rotation.

\subsection{How the Position Shift Impacts?}
\label{SEC3.3}
To quantitatively analyse how the position shift problem impacts the detector, we perform experiments on the relatively well aligned KAIST dataset by manually simulating the position shift.

\textbf{Baseline Detector} Following the setting in Adapted Faster R-CNN \cite{zhang2017citypersons}, we build the baseline detector and utilize RoIAlign \cite{he2017mask} to extract region features. We use the halfway fusion settings \cite{liu2016multispectral} for multispectral inputs.
On the KAIST dataset, this solid baseline detector achieves $15.2$ log-average miss rates (MR), about $11.0\%$ better than the $26.2$ MR reported in \cite{liu2016multispectral, li2019illumination}. 

\begin{figure}
\subfigure[Baseline]{
\begin{minipage}[t]{0.50\linewidth}
\centering
\includegraphics[width=1.56in]{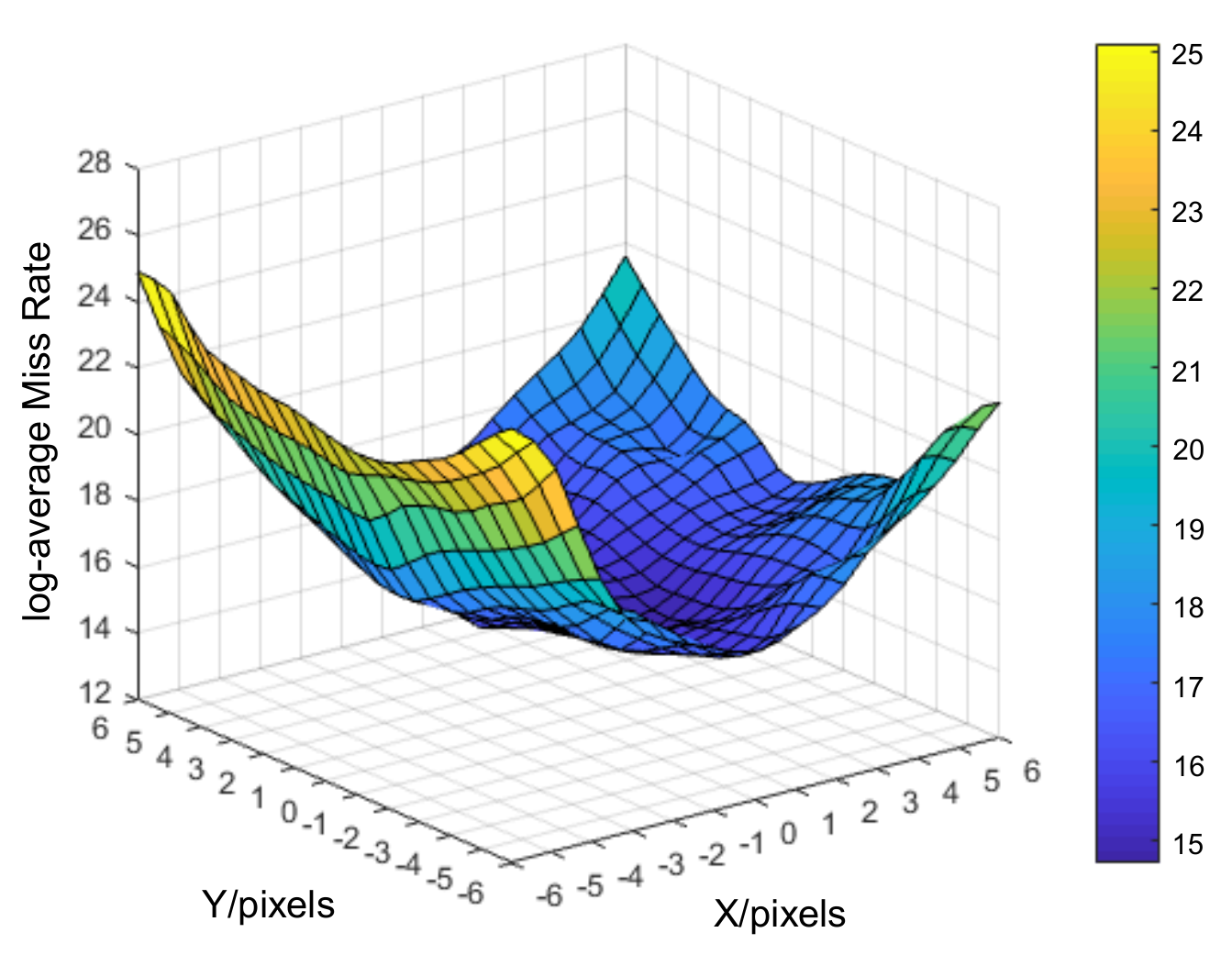}
%\caption{fig1}
\label{figure-shifting-1}
\end{minipage}}%
\subfigure[The proposed detector]{
\begin{minipage}[t]{0.50\linewidth}
\centering
\includegraphics[width=1.56in]{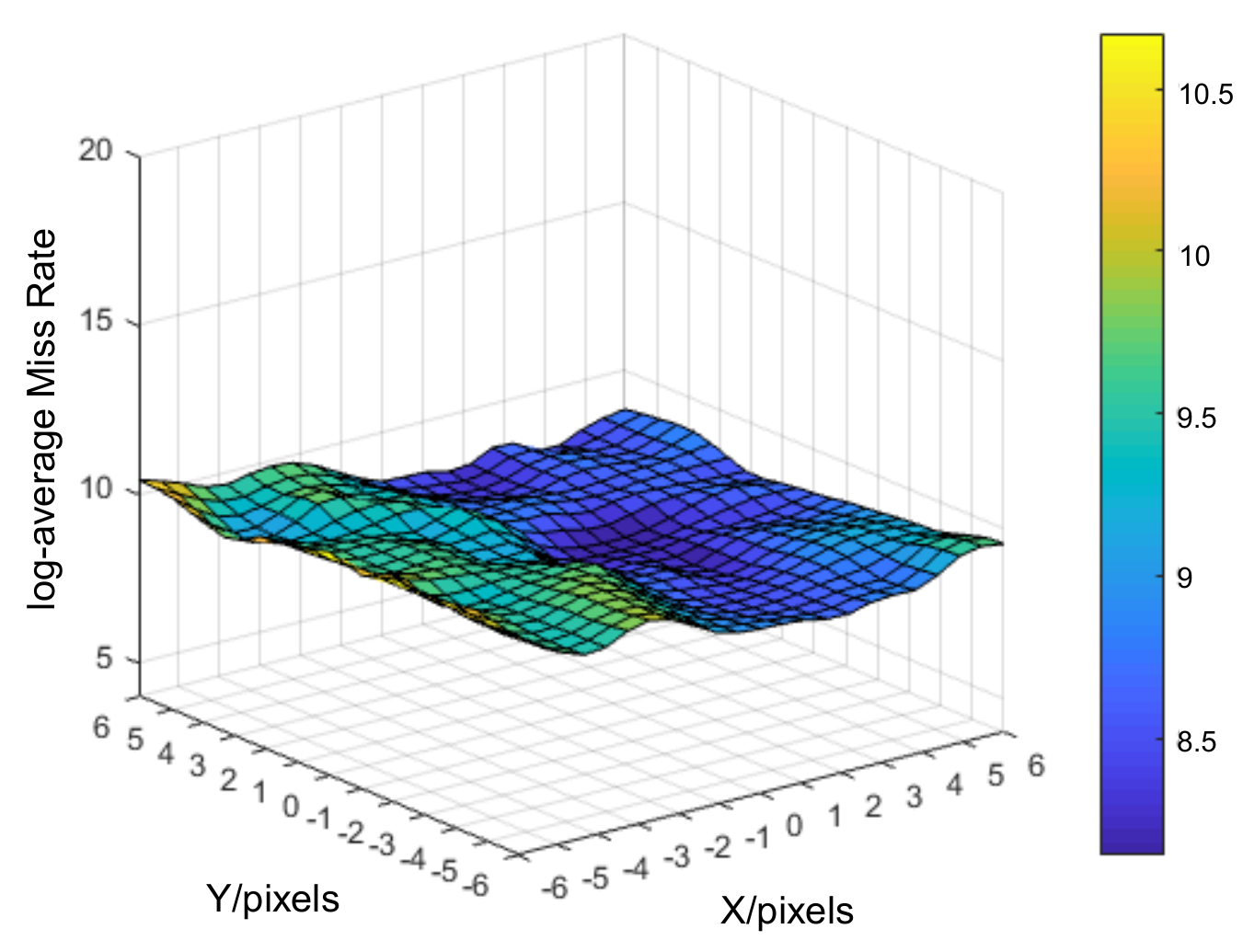}
%\caption{fig2}
\label{figure-shifting-2}
\end{minipage}}%
\caption{Surface plot of performances within the position shift experiments. The $x$ and $y$ axis denote different step sizes by which color images are shifted along. The vertical axis indicates the MR measured on the KAIST dataset, lower is better.} 
\label{figure-shifting}                     
\end{figure}

\textbf{Robustness to Position Shift} Without loss of generality, we set thermal images as the reference, hence the color images are spatially shifted along the $x$ and $y$ axis. 
We select $169$ different shift patterns in $\{(\Delta x,\Delta y)~|~\Delta x,\Delta y \in [-6,6];\Delta x,\Delta y \in Z \}$.
As illustrated in Figure \ref{figure-shifting-1}, the detection performance significantly drops as the absolute shift value increases. Especially, the worst-case $(\Delta x,\Delta y)=(-6,6)$ suffers about $\textbf{65\%}$ relative performance degradation (from $15.2$ to $25.1$ MR). The interesting thing is, the origin without any shift does not get the best performance. When we shift the color images to a specific direction $(\Delta x,\Delta y)=(1,-1)$, the detector achieves a better $14.7$ MR. This indicates that the detection performance can be further improved by appropriately tackling the position shift problem.

\begin{figure*}[!t]
\centering
\includegraphics[width=6.5in]{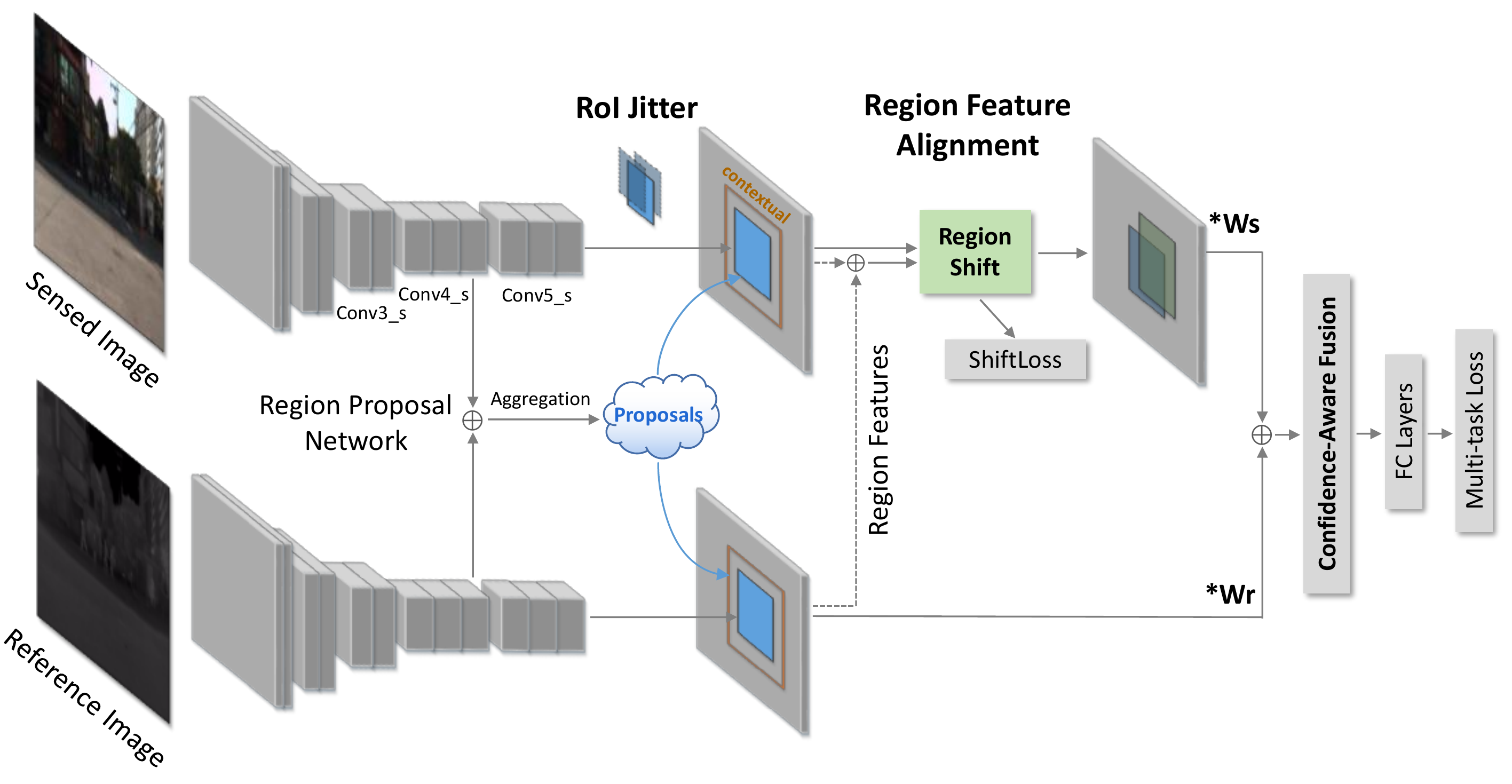}
\caption{The architecture of the proposed AR-CNN. We use color-thermal input as an example, after the two-stream feature extractor, numerous proposals are generated by the fused Region Proposal Network (RPN). Then the RFA module is utilized to predict and align the shift of regions, and an extra shift loss is calculated by utilizing the proposed KAIST-Paired annotation. Meanwhile, we perform the RoI Jitter strategy to enrich the shift patterns during training. 
After that, we can pool the aligned region features from the feature map of each modality. For the all-day RGB-T pedestrian detection, we utilize the confidence-aware fusion (CAF) method to pay attention to the more reliable modality, $W_{r}$ and $W_{s}$ denote the confidence weight of reference and sensed modality respectively. For the 3D RGB-D object detection, we calculate the multi-task loss based on the 3D box initialization and regression process. 
}
\label{figure-RFA}
\end{figure*}

%------------------------------------------------------------------------
\section{Our Method}
\label{S4}

In this section, we introduce the Aligned Region CNN (AR-CNN) framework and KAIST-Paired annotation (Section \ref{sec4.3.1}). The architecture is illustrated in Figure \ref{figure-RFA}, containing the region feature alignment module (Section \ref{sec4.1}), the RoI jitter training strategy (Section \ref{sec4.2}). We also present the confidence-aware fusion step (Section \ref{sec4.3.1}) for all-day RGB-T pedestrian detection and the adaption to RGB-D based 3D object detection task (Section \ref{sec4.4}). Algorithm \ref{alg:Framework} describes the whole pipeline.

\subsection{Region Feature Alignment}

To predict and align the shift between two modalities, we present the Region Feature Alignment (RFA) module in this subsection. In practice, the position shift is not a simple affine transformation due to the different properties of camera systems. As a result, the shift varies in different regions. It is usually large in regions near the image edge and relatively small near the image center. Therefore, the RFA module performs the shift prediction and feature alignment at the regional level.

\label{sec4.1}
\textbf{Reference and Sensed Modality} In image registration \cite{zitova2003image, brown1992survey}, the \textit{reference} and \textit{sensed} images refer to the stationary and transformed images, respectively. 
We introduce this concept into our multi-modal setting.
In the training phase, the reference modality is fixed and the feature alignment and RoI jitter strategy are performed on the sensed modality. 

\textbf{Proposals Generation} To keep more potential objectness regions, we aggregate both the reference and sensed feature maps (Conv4\_r and Conv4\_s) to generate the proposals. We utilize the Region Proposal Network (RPN) \cite{ren2015faster} for the proposal generation process.

\begin{figure}[tb]
\centering
\includegraphics[width=3in]{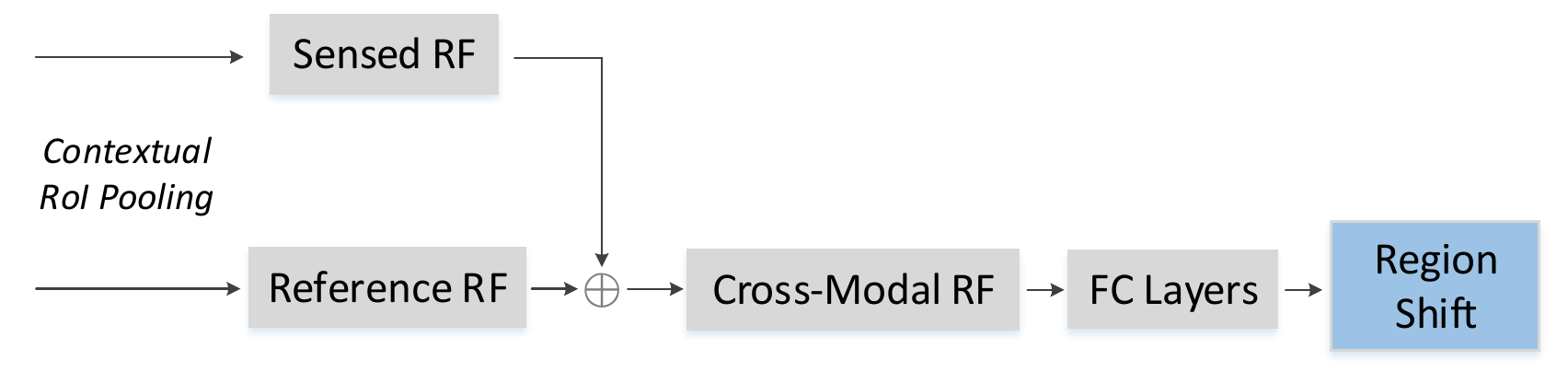}
% where an .eps filename suffix will be assumed under latex,
% and a .pdf suffix will be assumed for pdflatex; or what has been declared
% via \DeclareGraphicsExtensions.
\caption{The concrete structure of the proposed RFA module. Given the contextual region feature (RF) of each modality, this module obtains the cross-modal RF by conducting element-wise summation ($\oplus$). Then we utilize two FC layers to predict the position shift of each region of interest.}
\label{figure-RFA-1}
\end{figure}

\textbf{Feature Alignment} Figure \ref{figure-RFA-1} illustrates the concrete connection scheme of the RFA module. Specifically, the RFA module first enlarges the proposals (RoIs) to encompass the contextual information. Then we pool the contextual region features of each modality to small $H\times W$ (such as $7\times 7$) feature maps. Then the pooled region features are concatenated to obtain the cross-modal representation. Based on this representation, we use two consecutive fully connected layers to predict the shift (\ie $t_{x}$ and $t_{y}$) of each region. In this way, we obtain new coordinates of the sensed region and re-pool on this new region to get the aligned sensed features.

Since the proposed KAIST-Paired annotation (more details in Section \ref{sec4.3.1}) provides multi-modal bounding boxes, we can calculate the targets of ground truth shift as follow:
\begin{equation}
\begin{aligned}
&~t^{*}_{x} = (x_{s}-x_{r}) / w_r ~~~~~ t^{*}_{y} = (y_{s}-y_{r}) / h_{r}\\
\end{aligned}
\label{eqn-targets}
\end{equation}
where $x$ and $y$ are the center coordinates of the box, $w$ and $h$ refer to the width and height, $x_{r}$ and $x_{s}$ denote the reference and sensed ground truth box, $t^{*}_{x}$ and $t^{*}_{y}$ indicate the shift target for $x$ and $y$ coordinates, respectively.

\textbf{Adjacent Similarity Constraint} 
For natural images, the position shift is spatially smooth, \ie the shift targets for adjacent regions tend to be similar. Base on this, we add the adjacent similarity constraint to stabilize the training process of the RFA module. Specifically, we first randomly sample one of four nearest pixels (with feature stride) to the pooled features maps of the region of interest. Then, the same alignment targets are assigned to the region of interest and its neighbour, thus it encourages the RFA module to predict similar for adjacent regions.

\textbf{Position Shift Loss} To measure the accuracy of predicted shift targets, we calculate the position shift loss as follow:
%use the smooth L1 loss as the regression loss \cite{girshick2015fast}, \ie ,
\begin{equation}
\begin{aligned}
L_{shift}( \{p_{i}^{*}\}, \{t_{i}\}, \{t_{i}^{*}\}) = \\ \frac{1}{N_{shift}} \sum _{i=1} ^{n} p_{i}^{*} \mathrm{smoothL_{1}}(t_{i} - t_{i}^{*})
\end{aligned}
\end{equation}

The subscript $i$ denotes the index of RoIs in minibatch, $p^{*}_{i}$ is the class label ($1$ for the pedestrian and $0$ for the background) of the RoI, $t_{i}$ indicates the predicted shift, and $t^{*}_{i}$ is the ground truth shift target. $N_{shift}$ denotes the number of to-be-aligned RoIs. 

\begin{algorithm}[tb]
\caption{Framework of the multi-modal object detection system.}
\label{alg:Framework}
\begin{algorithmic}[1] 
\REQUIRE ~~\\ 
    The image of reference modality, $I_r$;\\
    The image of sensed modality, $I_s$;\\
    The output threshold $\tau$;
\ENSURE ~~\\ 
    A set of detection results $\mathcal{D}_r$ on reference images;\\
    \STATE Extracting the feature maps $Conv_r$ and $Conv_s$ from the reference and sensed modality;
    \STATE Aggregating $Conv_r$ and $Conv_s$ to generate a set of 2D proposals $\mathcal{P}$;
    \STATE Extracting region-wise features $\mathcal{F}_r$ and $\mathcal{F}_s$ of $\mathcal{P}$ by using RoI pooling;
    \STATE Predicting the position shift for $\mathcal{P}$ using the RFA module and re-extracting aligned region-wise features $\mathcal{F}_s^{'}$;
    \STATE Fusing region-wise features $\mathcal{F}_r$ and $\mathcal{F}_s^{'}$ with confidence-aware module to obtain $\mathcal{F}_f$;
    \STATE Conducting region-wise classification and (3D) bounding boxes regression to obtain confidences $C^i$ and coordinats $B^i$ for the $F_f^i \in \mathcal{F}_f$ of each proposal $P^i \in \mathcal{P}$;
    \STATE $\mathcal{D}_r \gets \varnothing$;
	\STATE \textbf{for} each $P^i \in \mathcal{P}$ \textbf{do}
	\STATE ~~~\textbf{if} $C^i > \tau$ \textbf{then}
	\STATE ~~~~~~$\mathcal{D}_r \gets \mathcal{D}_r \cup \{[B^i, C^i]\}$;
	\STATE ~~~\textbf{end if}
	\STATE \textbf{end for}
    %\STATE  $D_r = \{[B_0, C_0], [B_1, C_1]}, [B_n, C_n]\}$ 

\RETURN $\mathcal{D}_r$;

\end{algorithmic}
\end{algorithm}

\textbf{Multi-task Loss} Then the multi-task loss function is defined as follow:

\begin{equation}
\begin{aligned}
L(\{p_{i}\}, \{t_{i}\}, \{g_{i}\}, \{p_{i}^{*}\}, \{t_{i}^{*}\}, \{g_{i}^{*}\}) = L_{cls}(\{p_{i}\},\{p_{i}^{*}\})\\ + \lambda_{1} L_{shift}(\{p_{i}^{*}\}, \{t_{i}\}, \{t_{i}^{*}\}) + \lambda_{2} L_{asc}(\{p_{i}^{*}\}, \{\hat{t_{i}}\}, \{t_{i}^{*}\}) \\+ L_{reg}(\{p_{i}^{*}\}, \{g_{i}\}, \{g_{i}^{*}\})\\ 
\end{aligned}
\end{equation}

$L_{cls}$ and $L_{reg}$ are similar to the loss in Fast R-CNN \cite{girshick2015fast}. $L_{asc}$ is the loss of the adjacent similarity constraint, $\hat{t_{i}}$ is the predicted shift of the 4-neighbourhood. Variable $p_{i}$ refers to the predicted class confidence for the i-th RoI, $g_{i}$ is the predicted refinement, $p_{i}^{*}$ and $g_{i}^{*}$ are associated ground truths. We balance the $L_{shift}$ and $L_{reg}$ by setting the $\lambda$ as the weighting parameter. To weight the two terms, we use $\lambda_{1}=0.75, \lambda_{2}=0.25$. Apart from this multi-task loss, the definition of RPN loss follows the literature \cite{ren2015faster}.

\subsection{RoI Jitter Strategy}
\label{sec4.2}
\begin{figure}[t]
\subfigure[Reference modality]{
\begin{minipage}[t]{0.45\linewidth}
\centering
\includegraphics[width=1.48in]{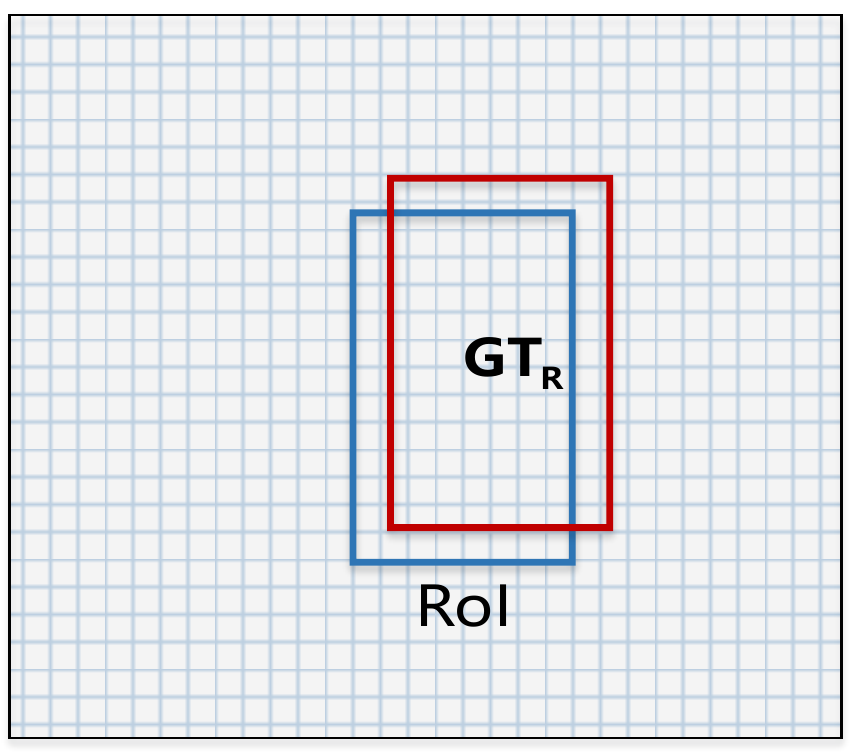}
%\caption{fig1}
\label{fig:roij:side:a}
\end{minipage}}%
\subfigure[Sensed modality]{
\begin{minipage}[t]{0.45\linewidth}
\centering
\includegraphics[width=1.48in]{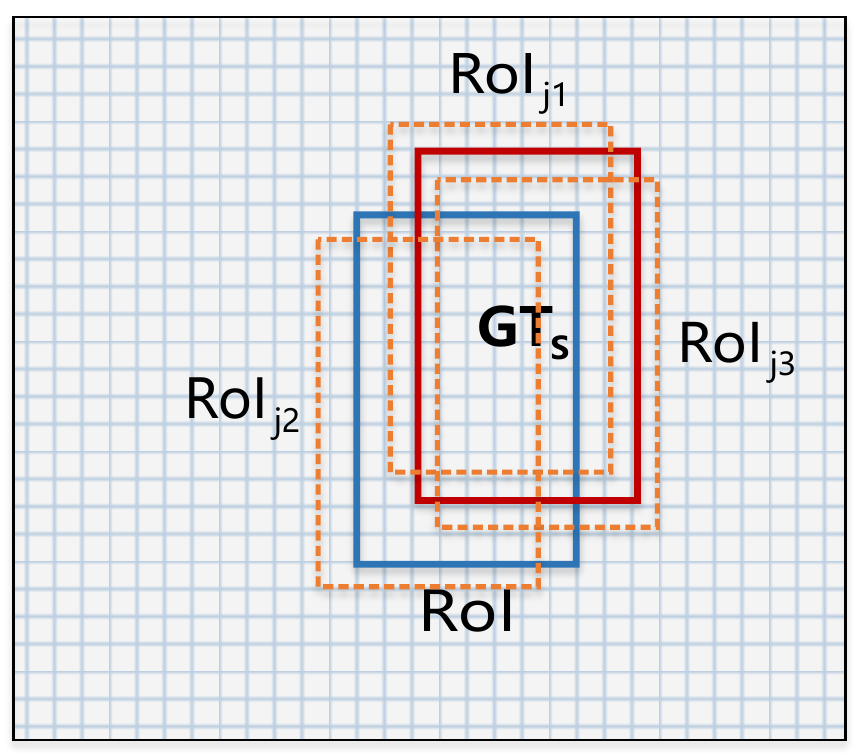}
%\caption{fig2}
\label{fig:roij:side:b}
\end{minipage}}%
\caption{Illustration of the proposed RoI jitter strategy. The red boxes are reference and sensed ground truth boxes ($\mathrm{GT}_{R}$ and $\mathrm{GT}_{S}$). The blue boxes stand for RoIs (proposals), which are shared by the two modalities. $\mathrm{RoI}_{j1,2,3}$ denote three potential proposals after jitter.} 
\label{figure-RoIJ}                     
\end{figure}

During training, the position shift can be limited to a certain kind of pattern, which is often related to the collection devices and settings of datasets. However, the practical shift pattern is unexpected. To fill the gap between the offline training and online testing, we present the RoI jitter strategy, as illustrated in Figure \ref{figure-RoIJ}. For the sensed modality, we introduce a random disturbance to generated new jittered positions of the proposal. Meanwhile, the shift targets of the RFA module are enriched correspondingly. The random jitter target is derived from the Gaussian distribution as follow:
\begin{equation}
\begin{aligned}
t^{j}_{x},~ t^{j}_{y} \sim N(0,\sigma_{0}^{2};&0,\sigma_{1}^{2};0)\\
\end{aligned}
\end{equation}

Variable $t^{j}$ indicates jitter targets along the $x$ and $y$ axis, $\sigma$ denotes the radiation extent of jitter.
Afterward, we use the inverse process of Equation \ref{eqn-targets} to jitter the $\mathrm{RoI}$ to $\mathrm{RoI}_{j}$.

\textbf{Minibatch Sampling}
To select the set of samples for minibatch training, we need to define the positive and negative samples by calculating the IoU overlap.
In our setting, we consistently use the reference RoI for this calculation, since we perform the random jitter on the sensed RoI.

\subsection{All-day RGB-T Pedestrian Detection}
\label{sec4.3}

\subsubsection{KAIST-Paired Annotation}
\label{sec4.3.1}
Existing RGB-T datasets ignore the cross-modal correlation between pedestrians. To fill this gap, we provide the multi-modal labelling following these rules:
\begin{itemize}
\item{Multi-modal locating. We locate the pedestrian in both color and thermal modality, thus providing clear position shift patterns.}
\item{Adding relationships. To indicate the cross-modal correlations of objects, we assign unique indexes to the pedestrians.}
\item{Annotating unpaired pedestrians. If a pedestrian only appears in a single modality, we annotate it as the ``unpaired'' one.}

\end{itemize}

\textbf{Statistics of KAIST-Paired}
Since we have access to the proposed KAIST-Paired annotation, the statistics of position shift in the original KAIST dataset can be derived.  As shown in Figure \ref{figure-stat-shifting-1}, over half of pedestrians suffer from the position shift, which mostly ranges from $0$ to $10$ pixels.

\begin{figure}[t]
\centering
\includegraphics[width=3.3in]{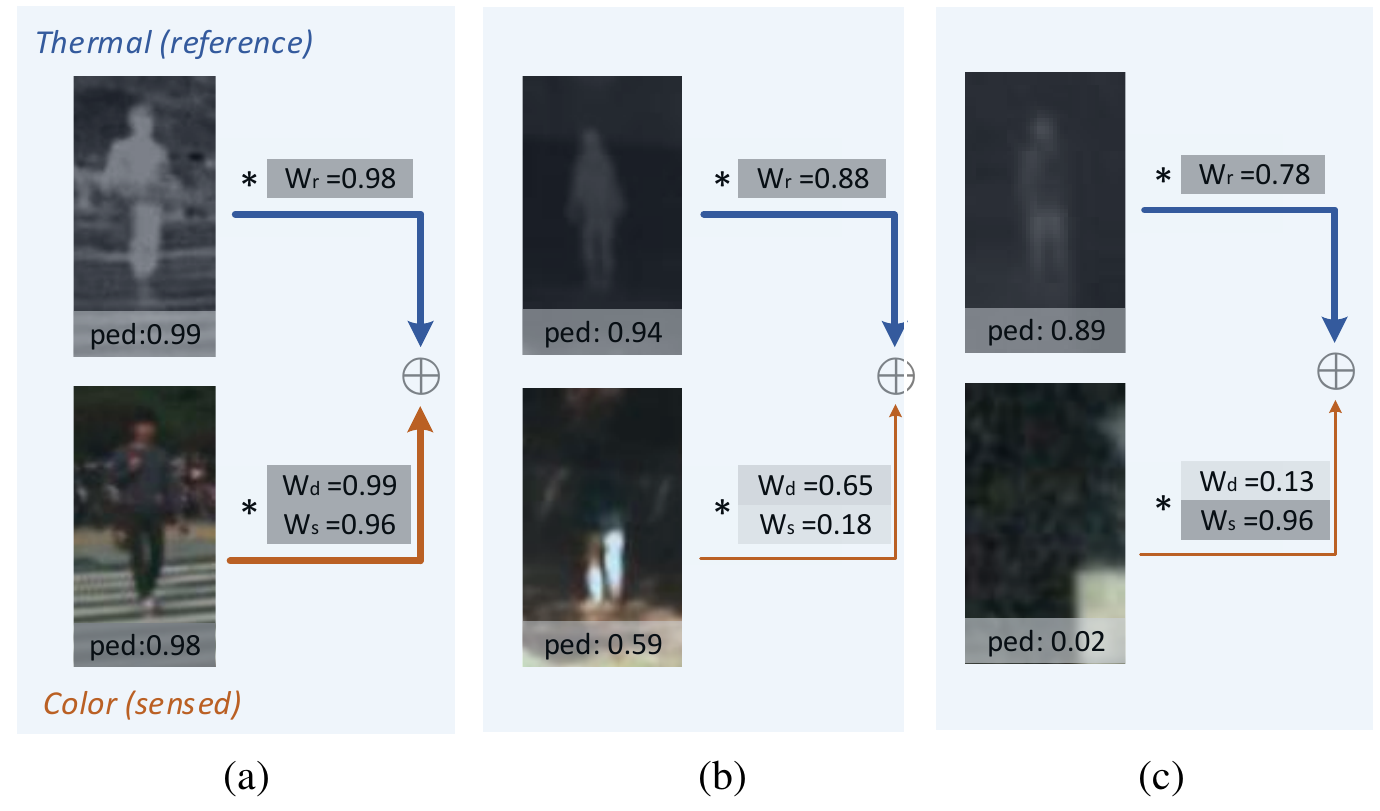}
% where an .eps filename suffix will be assumed under latex,
% and a .pdf suffix will be assumed for pdflatex; or what has been declared
% via \DeclareGraphicsExtensions.
\caption{The confidence-aware fusion module. We illustrate three typical situations in RGB-T pedestrian detection: (a) in the daytime, both modalities present clear imaging; (b) when the illumination is bad, the pedestrian in color modality is difficult to distinguish, hence more attention will be paid to the thermal modality; (c) the pedestrian is unpaired, only existing in a single modality, hence the color feature will be depressed.}
\label{figure-CAF}
\end{figure}

\subsubsection{Confidence-Aware Fusion}
\label{sec4.3.2}
In all-day operations, the quality of information from color and thermal modality is diverse: features from the color modality are distinguishable in the daytime yet fade in the nighttime; and thermal images provide the pedestrian silhouette around the clock but lacks visual details like the clothing. 
To make full use of the characteristics between two different modalities, we propose the confidence-aware fusion method to select the more reliable feature while suppressing the less useful one via feature re-weighting. 

As illustrated in Figure \ref{figure-CAF}, the confidence-aware fusion module uses two confidence weights $\mathrm{W}_{r}$ and $\mathrm{W}_{f}$.
To calculate the two weights, we add a branch for each modality to obtain classification score $p$. Then the confidence weights for each modality are calculated as: $\mathrm{W}_{r} = |p^{1}_{r}-p^{0}_{r}|$, $\mathrm{W}_{s} = |p^{1}_{s}-p^{0}_{s}|$, where $p^{1}$ and $p^{0}$ are the probability of foreground and background, $r$ denotes the reference modality and $s$ is the sensed one. 
Then this module performs the multiplication to pay more attention to the reliable modality.

\textbf{Unpaired Objects} To mitigate the ambiguous classification for unpaired objects, we use the disagreement weight, $\mathrm{W}_{d} = 1-|p^{1}_{r} - p^{1}_{s}| = 1-|p^{0}_{r} - p^{0}_{s}|$. If the sensed modality provides  a contradictory prediction with the reference, its feature will be suppressed.

%%% TABLE-CVC

\begin{table}[t]
\begin{center}
\linespread{1.2}\selectfont
\begin{tabular}{c|c|c|c|c}
\toprule 
\multirow{2}*{~}&\multirow{2}*{Method} & \multicolumn{3}{c}{MR} \\
\cline{3-5}
&&Day&Night&All\\

\hline
\multirow{5}*{\rotatebox{90}{Visible}}&\multirow{1}*{SVM \cite{gonzalez2016pedestrian}}&~37.6~&~76.9~&~-~\\

&\multirow{1}*{DPM \cite{gonzalez2016pedestrian}}&~25.2~&~76.4~&~-~\\

&\multirow{1}*{Random Forest \cite{gonzalez2016pedestrian}}&~26.6~&~81.2&~-~\\

&\multirow{1}*{ACF \cite{Park2018Unified}}&~65.0~&~83.2&~71.3~\\

&\multirow{1}*{Faster R-CNN \cite{Park2018Unified}}&~43.2~&~71.4~&~51.9~\\

\hline

\multirow{5}*{\rotatebox{90}{Visible+Thermal~}}&\multirow{1}*{MACF \cite{Park2018Unified}}&~61.3~&~48.2~&~60.1~\\

&\multirow{1}*{Choi \etal} \cite{choi2016multi} &~49.3~&~43.8~&~47.3~\\

&\multirow{1}*{Halfway Fusion \cite{Park2018Unified}}&~38.1~&~34.4~&~37.0~\\

&\multirow{1}*{Park \etal \cite{Park2018Unified}} &~31.8~&~30.8~&~31.4~\\

\cline{2-5}

&\multirow{1}*{AR-CNN (Ours)}&~\textbf{24.3}~&~\textbf{18.1}~&~\textbf{22.0}~\\
\bottomrule
\end{tabular}
\end{center}
\caption{Detection performances on the CVC-14 dataset. The first column denotes the input modalities of the method. For ACF, MACF, Faster R-CNN, and Halfway Fusion, we use the re-implementation in the literature \cite{Park2018Unified}.}
\label{table-cvc}
\end{table}

%%% TABLE-drone
\begin{table}[t]
\begin{center}
\linespread{1.2}\selectfont
\begin{tabular}{c|c|c|c|c|c}
\toprule  
\multirow{1}*{Method} &mAP&car&bus&truck&cyclist \\

\hline
\multirow{1}*{SSD\cite{liu2016ssd}}&23.6&40.6&31.7&21.3&6.8\\

\multirow{1}*{Faster R-CNN\cite{ren2015faster}}&26.7&44.7&30.7&21.9&9.5\\

\hline

\multirow{1}*{Halfway Fusion\cite{liu2016multispectral}}&31.9&52.6&37.7&25.0&12.4\\

\multirow{1}*{CIAN\cite{zhang2019cross}}&34.8&57.0&40.2&28.7&13.3\\

\multirow{1}*{AR-CNN (Ours)}&\textbf{36.4}&\textbf{59.8}&\textbf{40.6}&\textbf{29.9}&\textbf{15.1}\\
\bottomrule
\end{tabular}
\end{center}
\caption{Object detection results on the drone-based RGB-T dataset. Only RGB images are used to train the SSD and Faster R-CNN. RGB-T image pairs are used to train the Halfway Fusion, CIAN, and AR-CNN model. We use mAP to evaluate the performance of detectors.}
\label{table-drone}
\end{table}

\subsection{3D RGB-D Object Detection}
\label{sec4.4}
To better understand the physical 3D world, object detection also aims to predict the 3D location and its full extent in 3D space. In this paper, we have RGB-D image pairs as inputs and leverage the 2.5D representation to predict 3D bounding boxes.

\textbf{2D RoI proposals} Since RGB and depth images contain complementary information, we use both modalities to generate the proposals in 2D space. As in Section \ref{sec4.1}, the Region Proposal Network (RPN) is adapted to generate 2D proposals in an end-to-end fashion.

\textbf{3D bounding box initialization and regression} 
Given several 2D proposals, we use some basic transformations to initialize the related 3D bounding boxes. Following the settings in \cite{deng2017amodal}, each 3D bounding box is parametrized
into one seven-entry vector $[x_{cam} , y_{cam} , z_{cam} , l, w, h, \theta]$. $[x_{cam} , y_{cam} , z_{cam}]$ is the centroid under camera coordinate system, and $[l, w, h]$ is the 3D dimension. The parameter $\theta\in[-\pi/2,\pi/2]$ is the angle between the principal axis and its orientation vector under the tilt coordinate system, in which the point clouds are aligned with gravity direction. Hence this rotation is only around the gravity direction.

For initialization, the median of depth value in each individual region is used as the initialization of $z$ axis, noted as $z_{ini}$, then the $x_{ini}$ and $y_{ini}$ can be calculated as follow:
\begin{equation}
\left\{
\begin{aligned}
x_{ini} &= z_{ini} * (c_{x}-o_{x}) / f \\
y_{ini} &= z_{ini} * (c_{y}-o_{y}) / f \\
\end{aligned}
\right.
\end{equation}
where $f$ is the focal length of RGB camera, $(o_{x}, o_{y})$ is the
principal point, $(c_{x}, c_{y})$ is the center coordinate of 2D box proposal. The 3D size $[l, w, h]$ is set as the class-wise average dimension, which is inspired by the \textit{familiar size} in human 3D perception \cite{fredebon1992role, kar2015amodal, deng2017amodal}. The angle $\theta$ is set to $0$ by default.

\begin{center}
\begin{figure*}[htbp]
\centering
\subfigure[Day, MR]{                    
\begin{minipage}{0.3\linewidth}
   \centering                                                       
\includegraphics[scale=0.36]{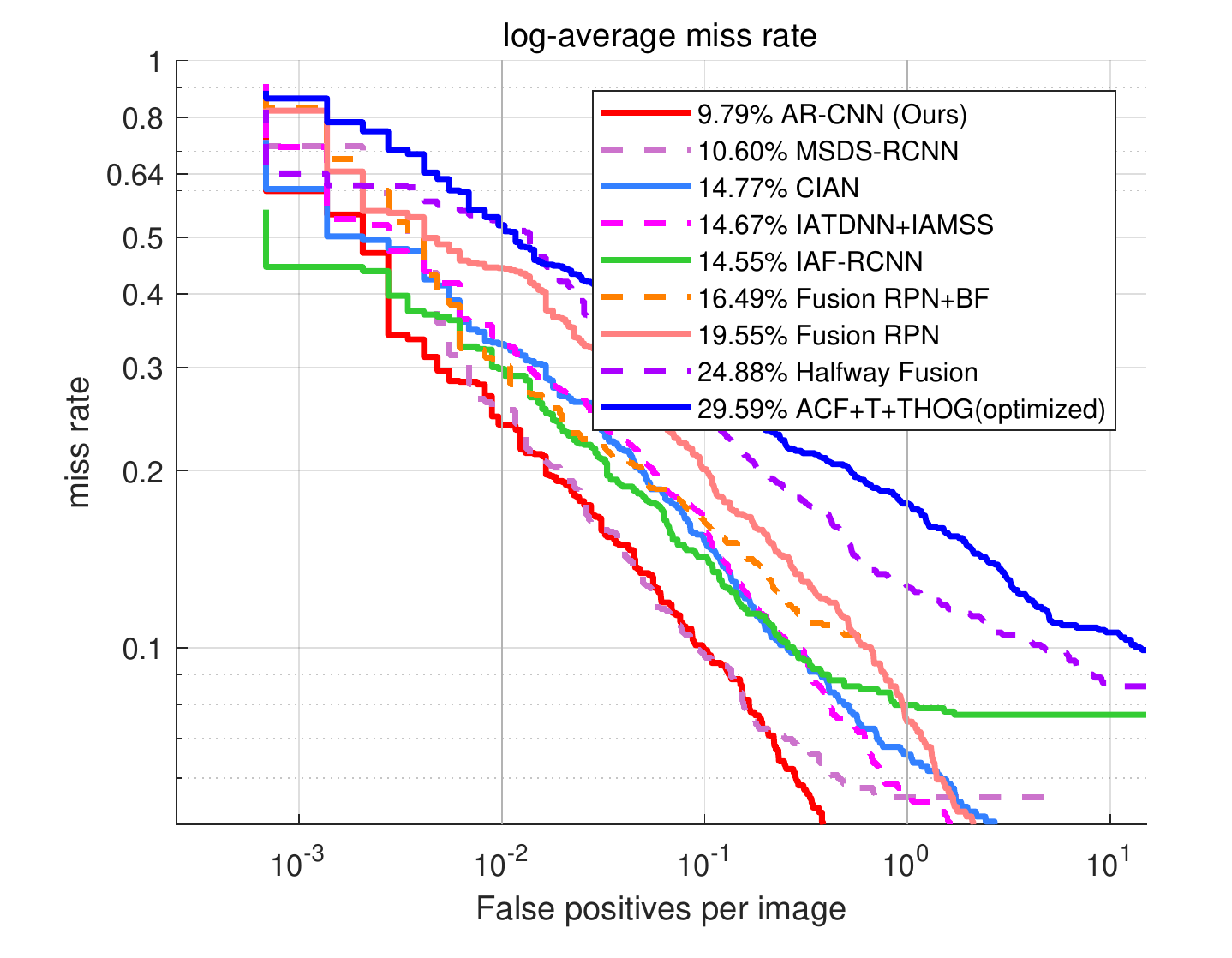}                
\end{minipage}}
\subfigure[Night, MR]{                    
\begin{minipage}{0.31\linewidth}
\centering                                                          
\includegraphics[scale=0.36]{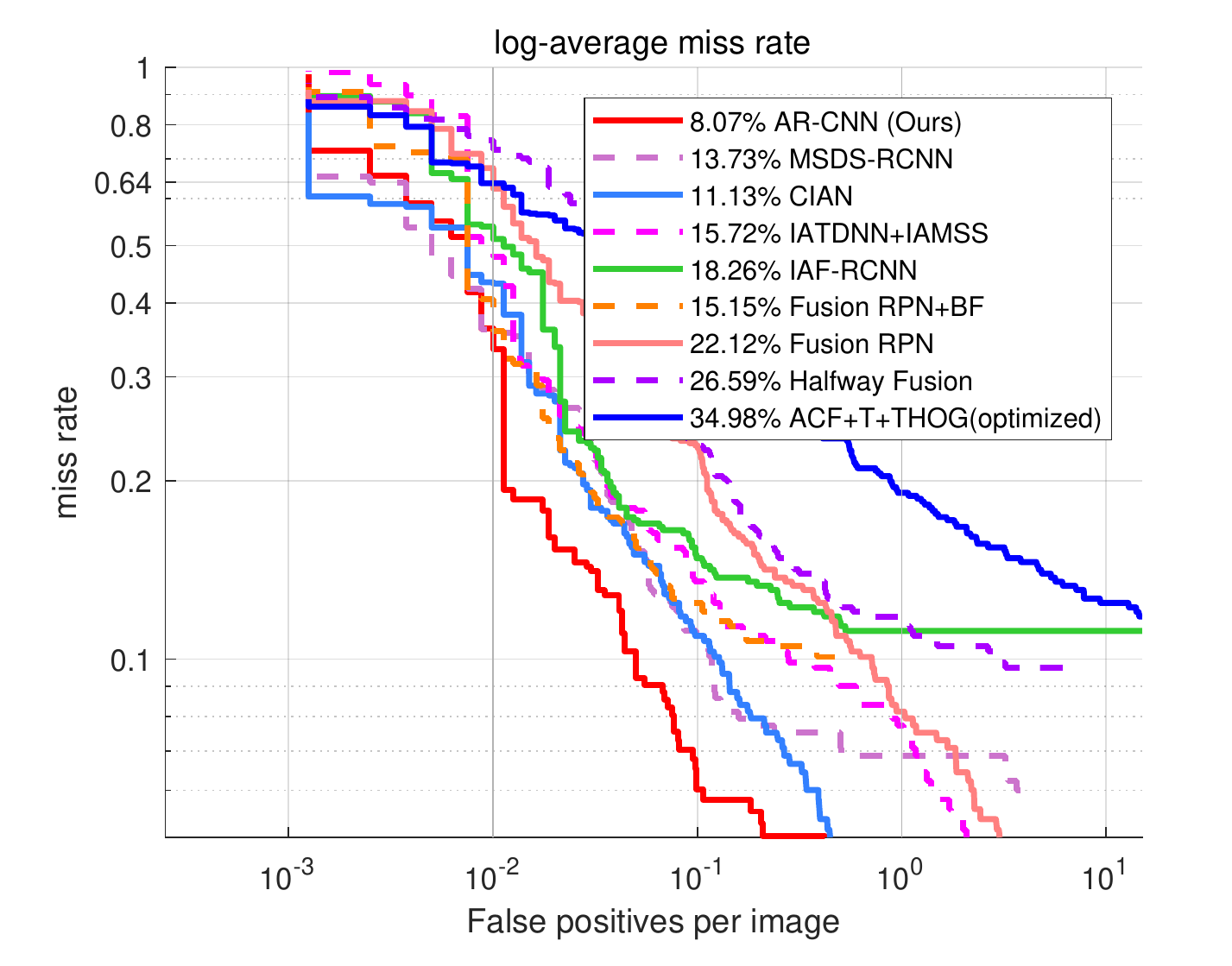}                
\end{minipage}}
\subfigure[All-day, MR]{                    
\begin{minipage}{0.31\linewidth}
\centering                                                          
\includegraphics[scale=0.36]{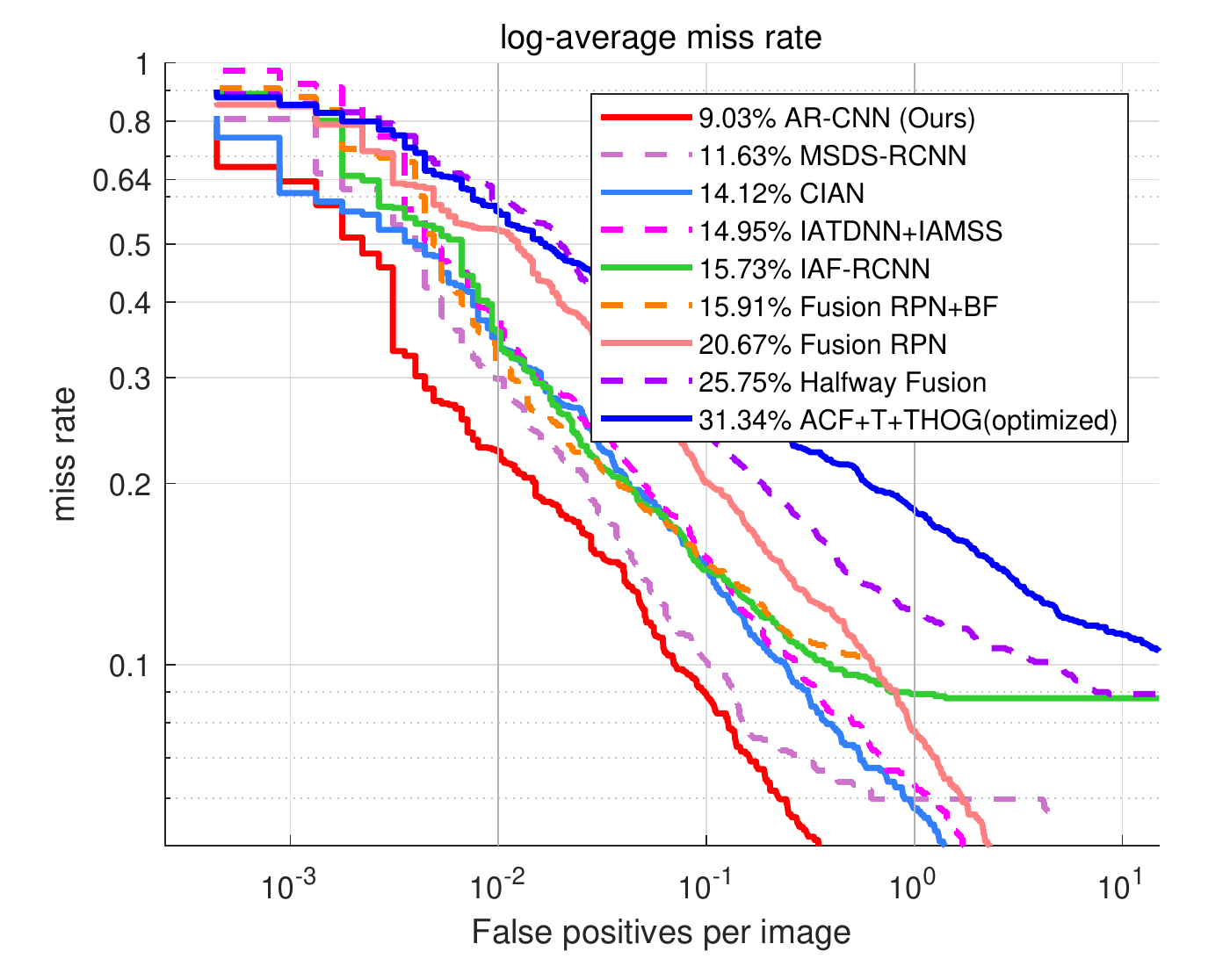}                
\end{minipage}}

\subfigure[Day, MR$^{C}$]{                    
\begin{minipage}{0.31\linewidth}
   \centering                                                       
\includegraphics[scale=0.36]{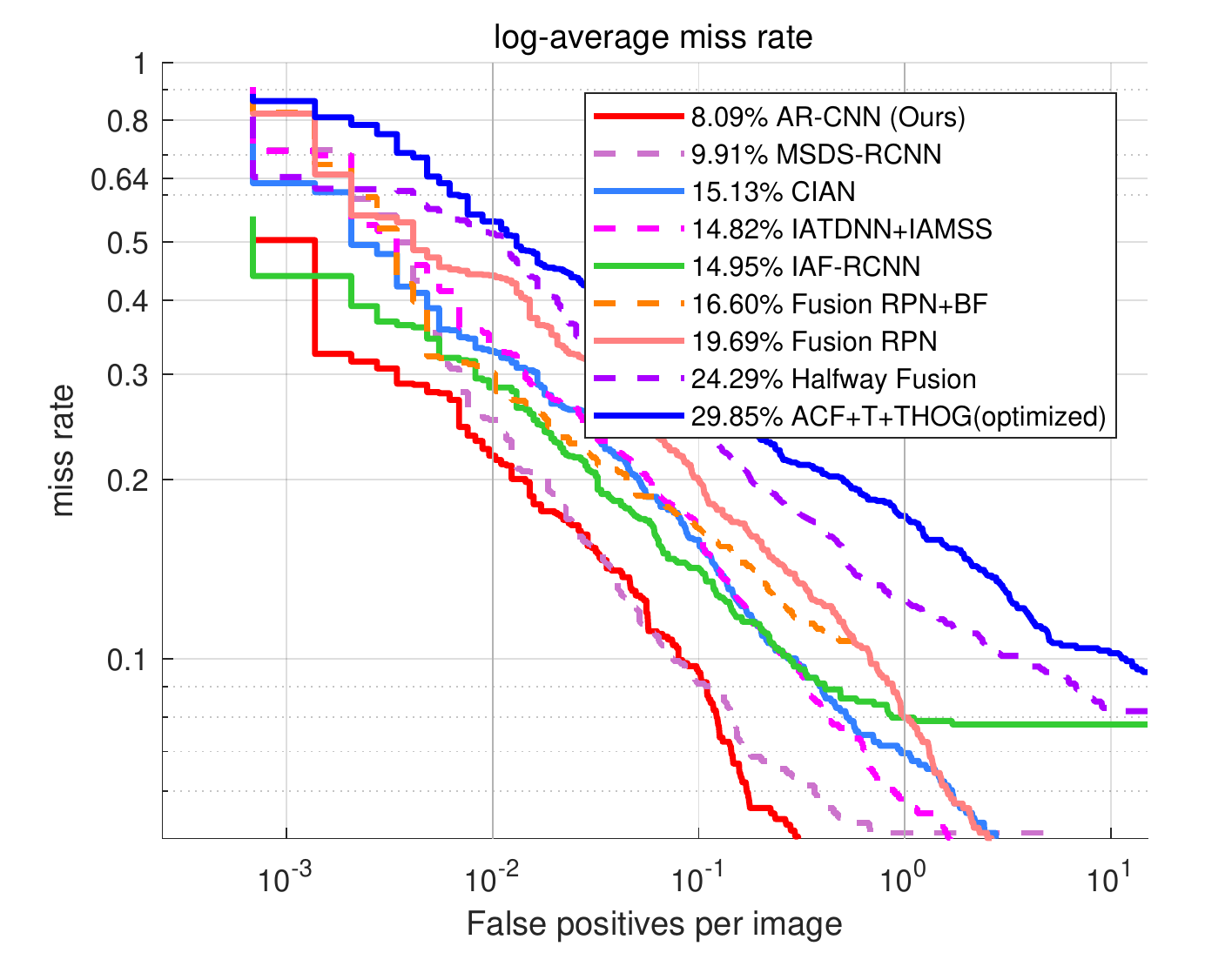}                
\end{minipage}}
\subfigure[Night, MR$^{C}$]{                    
\begin{minipage}{0.31\linewidth}
\centering                                                          
\includegraphics[scale=0.36]{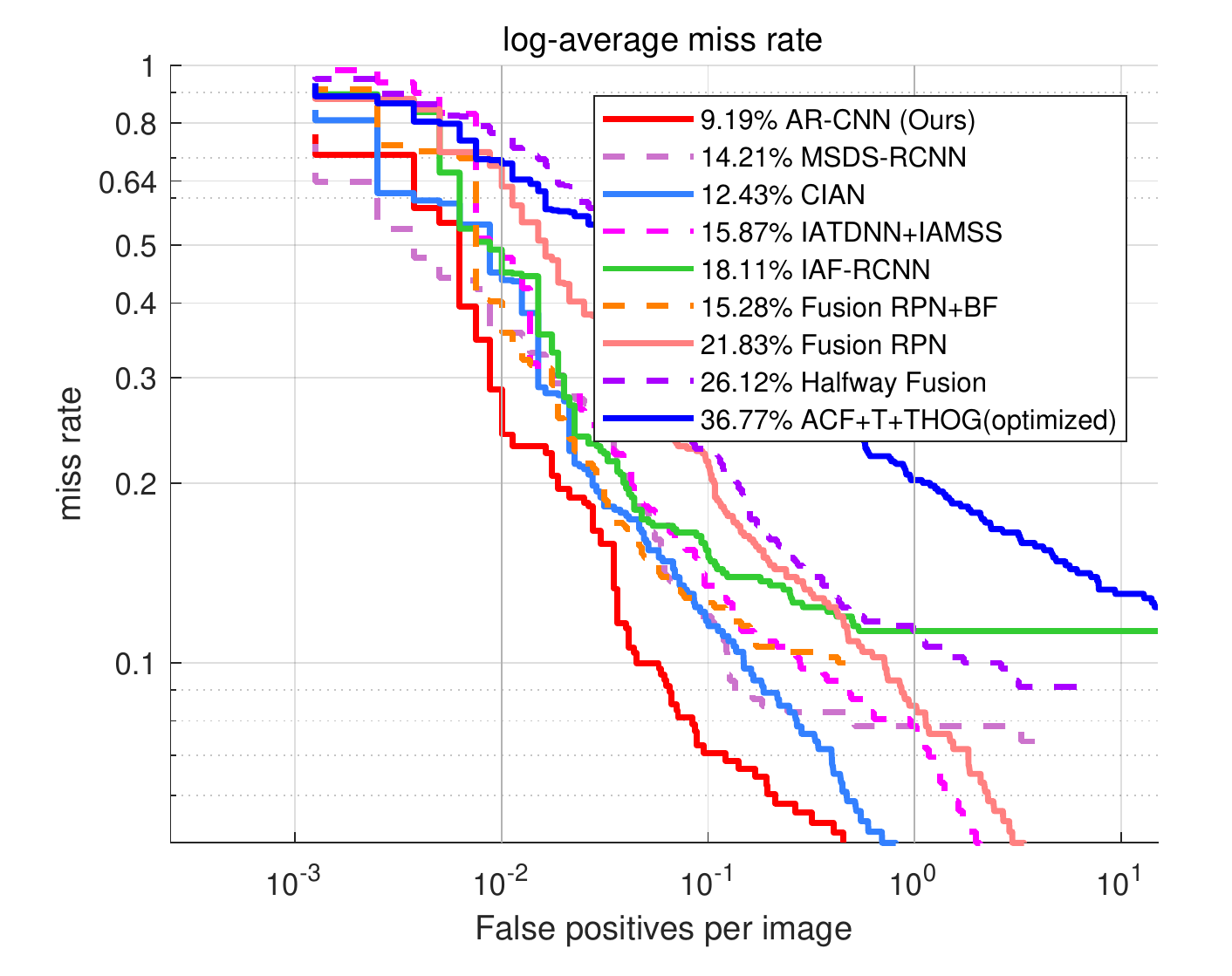}                
\end{minipage}}
\subfigure[All-day, MR$^{C}$]{                    
\begin{minipage}{0.31\linewidth}
\centering                                                          
\includegraphics[scale=0.36]{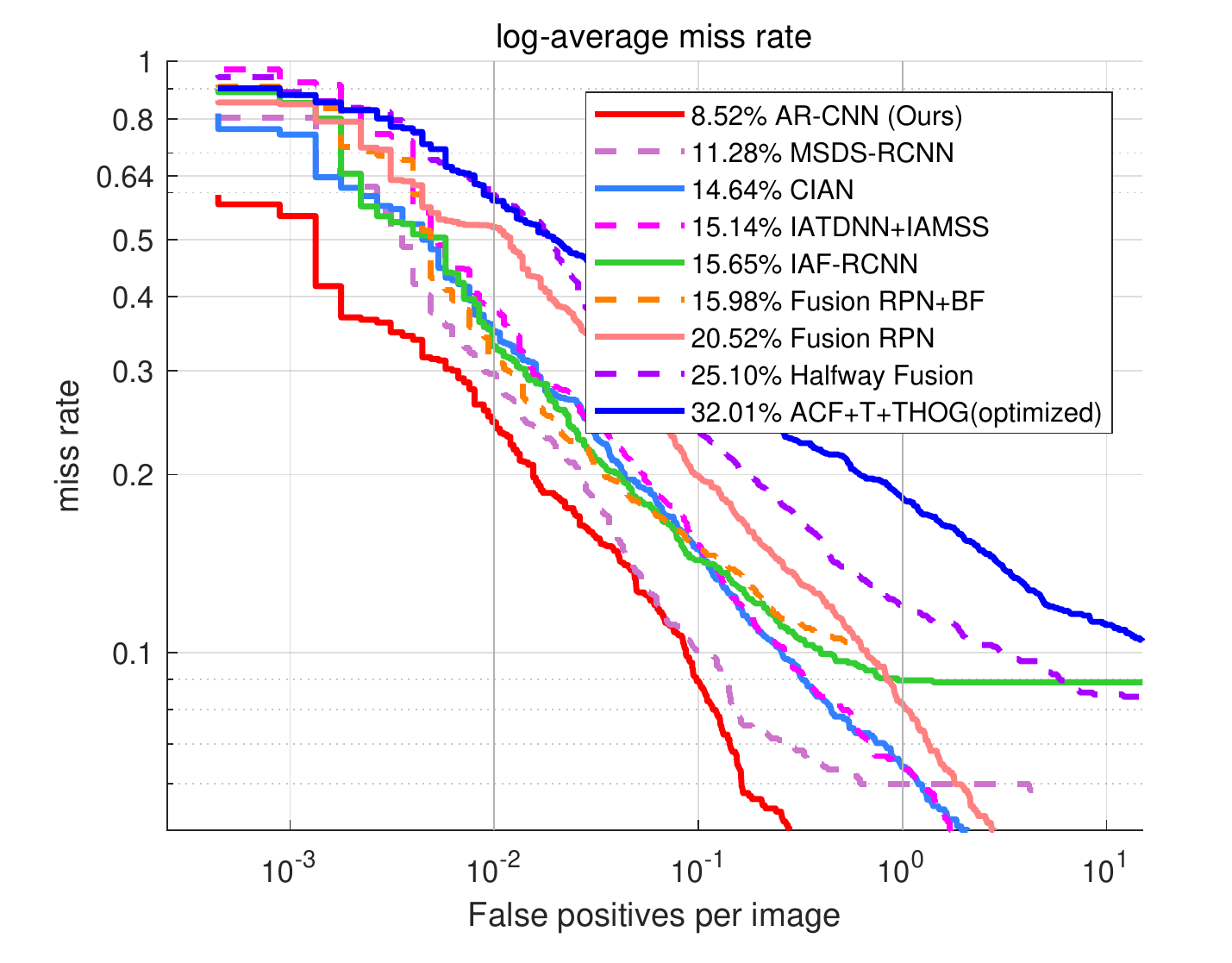}                
\end{minipage}}

\subfigure[Day, MR$^{T}$]{                    
\begin{minipage}{0.31\linewidth}
   \centering                                                       
\includegraphics[scale=0.36]{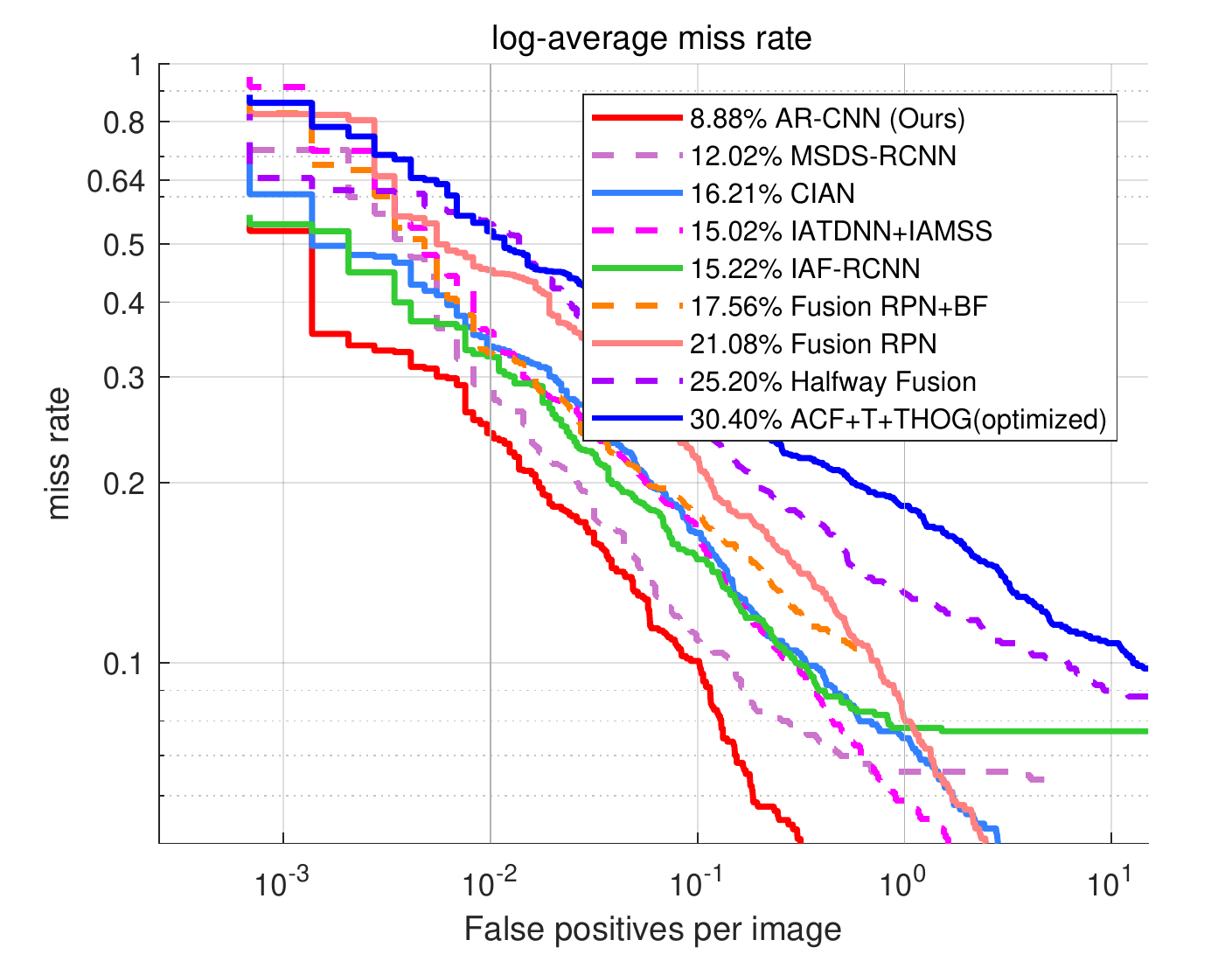}                
\end{minipage}}
\subfigure[Night, MR$^{T}$]{                    
\begin{minipage}{0.31\linewidth}
\centering                                                          
\includegraphics[scale=0.36]{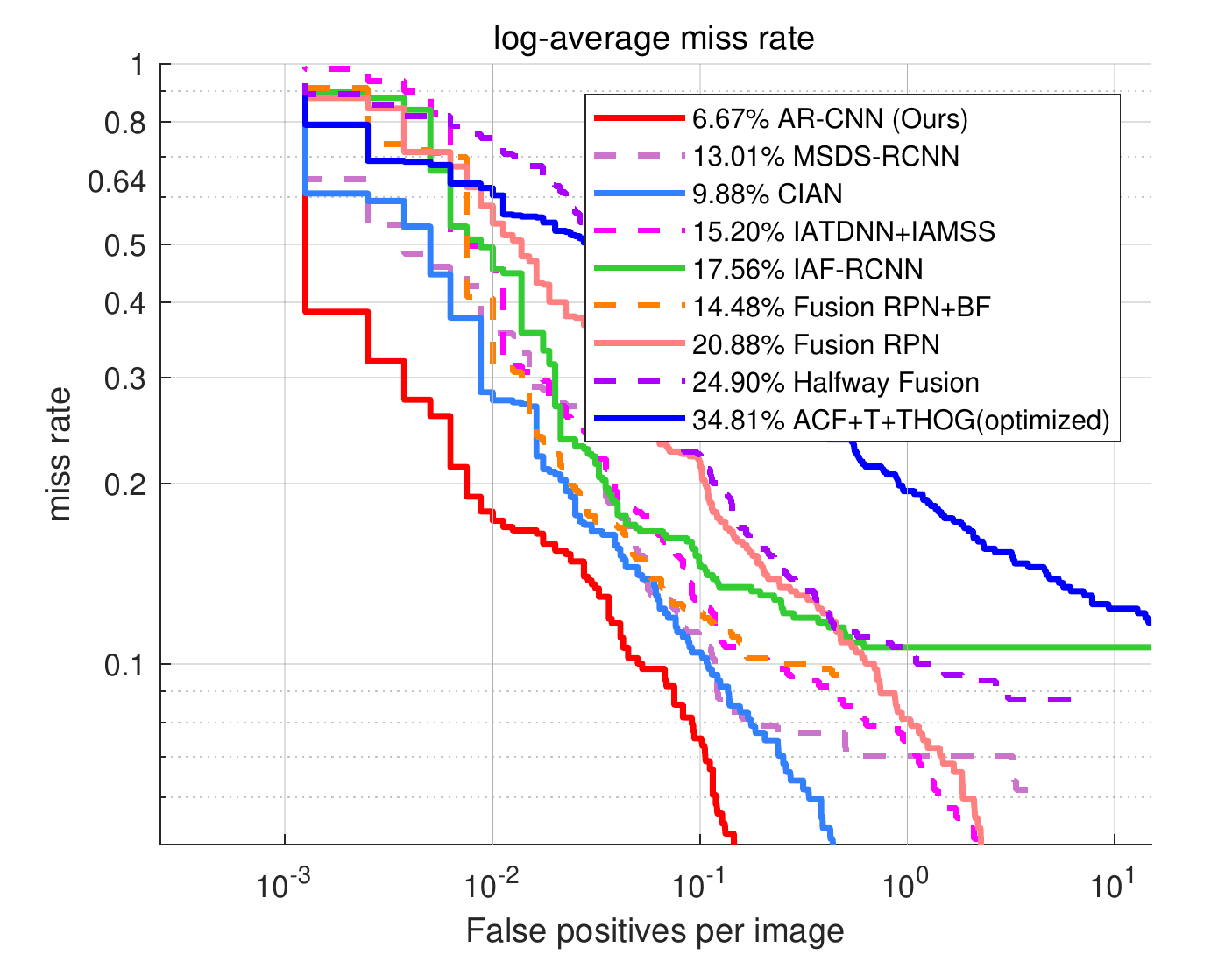}                
\end{minipage}}
\subfigure[All-day, MR$^{T}$]{                    
\begin{minipage}{0.31\linewidth}
\centering                                                          
\includegraphics[scale=0.36]{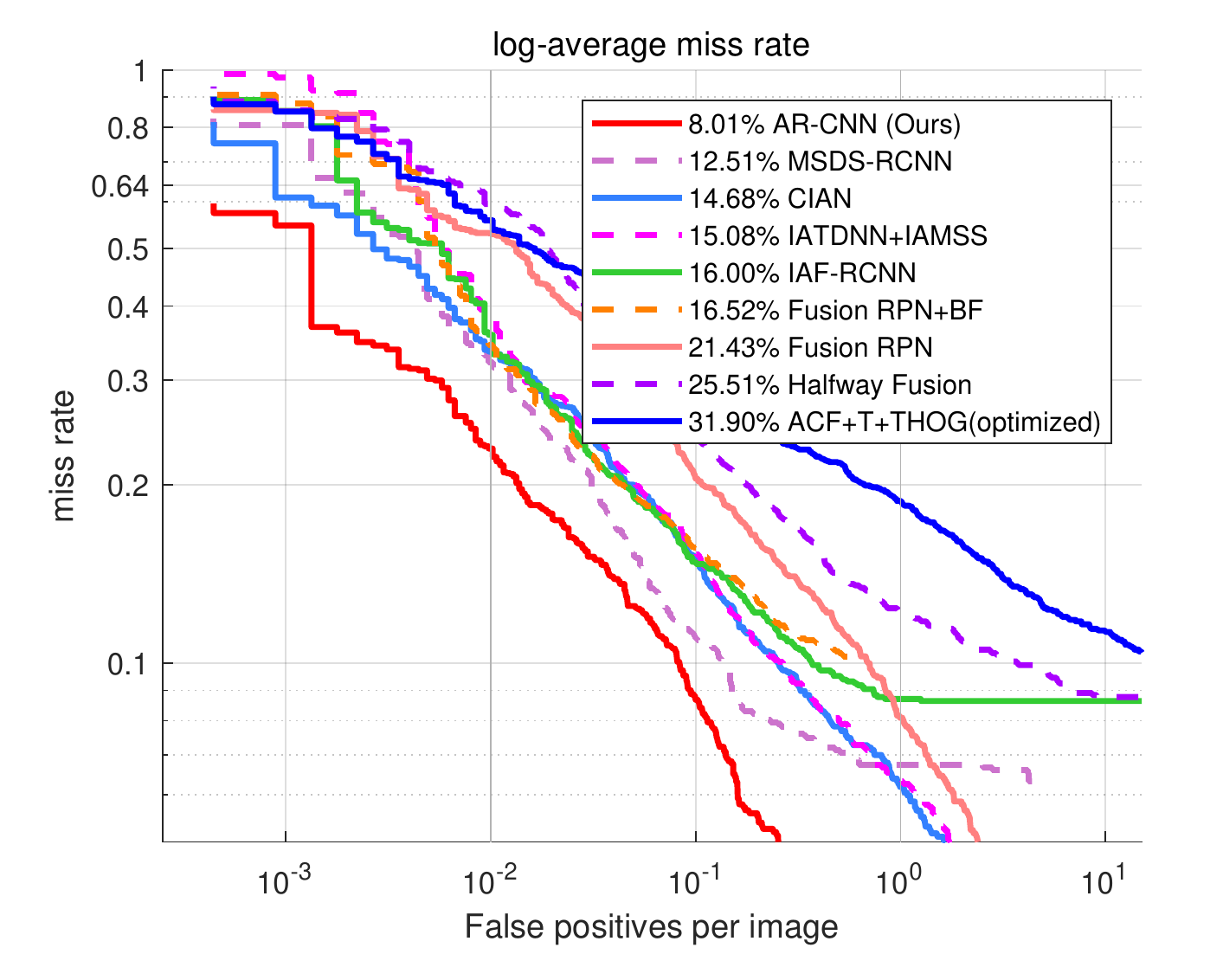}                
\end{minipage}}

\caption{Comparisons of the miss rate curves reported on the KAIST dataset. The performance is evaluated by the MR protocol, and also MR$^{C}$ and MR$^{T}$ using the KAIST-Paired annotation. }                       
\label{figure-MR}                                                        
\end{figure*}
\end{center}

With the seven-entry targets prediction between initialized vector and targeted vector, \ie $[v_{x} , v_{y} , v_{z} , v_{l}, v_{w}, v_{h}, v_{\theta}]$, the 3D regeression loss can be calculated as follow: 
\begin{equation}
\begin{aligned}
L_{3dreg}(p^{*}, v_{3d}, v_{3d}^{*})= [p^{*}>=1]~\mathrm{smoothL_{1}}(v^{*}_{3d}- v_{3d})\\ 
\end{aligned}
\end{equation}
where $p^{*}$ is the ground-truth class index, $v^{*}_{3d}$ are the ground truth regression offsets, and $v_{3d}$ are the predicted regression targets.

%------------------------------------------------------------------------
\section{Experiments on Case 1: RGB-T Object Detection}
\label{S5}

In this section, we first conduct several experiments on the challenging KAIST \cite{hwang2015multispectral} and CVC-14 \cite{gonzalez2016pedestrian} RGB-T pedestrian datasets. We evaluate all methods on the ``reasonable'' setup in which pedestrians are not or partially occluded and the heights are larger than 55 pixels. Then, to verify the generalization of our method on object detection, we collect a drone-based RGB-T dataset which includes 4 categories of objects: car, truck, bus, cyclist.  

\subsection{Dataset}
\textbf{KAIST} KAIST multispectral pedestrian dataset \cite{hwang2015multispectral} includes $95,328$ color-thermal frame pairs with $103,128$ annotations. To cover diverse illumination conditions, the dataset is captured in different scenes, through day and night. For the testing, we use the test set containing $2,252$ image pairs sampled from the video with $20$-frame skips.

\textbf{CVC-14} CVC-14 dataset \cite{gonzalez2016pedestrian} uses both visible and FIR cameras to record color-thermal video sequences during day and night time. It contains $8,518$ frames, among which $7,085$ frames are for training and $1,433$ for testing. The two cameras are not in a well-calibrated condition. In consequence, this dataset suffers more from the position shift problem, as shown in Figure \ref{figure-stat-shifting-2}. This problem makes it difficult for state-of-the-art multi-modal detectors to apply to the CVC-14 dataset \cite{xu2017learning, li2019illumination, guan2019fusion, zhang2019cross, cao2019box}.

\begin{table*}[!htbp]
\begin{center}
\linespread{1.18}\selectfont
\begin{tabular}{cccc|c|c|c|c|c|c|c|c|c}
\toprule  
\multicolumn{4}{c|}{\multirow{2}*{Method}} &  \multicolumn{3}{c|}{$S^{0^{\circ}}$} & \multicolumn{2}{c|}{$S^{45^{\circ}}$} & \multicolumn{2}{c|}{$S^{90^{\circ}}$} & \multicolumn{2}{c}{$S^{135^{\circ}}$}\\
\cline{5-13}
&&&&$O$ &$\mu$ &$\sigma$ &$\mu$ &$\sigma$  &$\mu$ &$\sigma$ &$\mu$ &$\sigma$\\

\hline

\multicolumn{4}{c|}{Halfway Fusion \cite{liu2016multispectral}}&~25.51~ &~35.72~ &~8.14~ &~35.19~ &~8.68~ &~34.07~ &~7.95~ &~34.59~ &~8.87~\\

\multicolumn{4}{c|}{Fusion RPN \cite{konig2017fully}}&~21.43~ &~31.57~ &~8.34~ &~30.42~ &~9.97~ &~29.96~ &~7.02~ &~33.14~ &~9.85~\\%&~21.43~ &~37.52~ &~15.41~ &~26.6~ &~46.2~ &~22.4~ &~1.5~ &~26.0~ &~22.6~\\

\multicolumn{4}{c|}{Adapted Halfway Fusion}&~15.59~ &~ 25.02 ~ &~8.29~ &~25.85~ &~10.26~ &~22.85~ &~7.53~ &~26.37~ &~10.14~\\

\multicolumn{4}{c|}{CIAN \cite{zhang2019cross}}&~14.68~ &~24.50~ &~8.51~ &~22.97~ &~10.70~ &~20.01~ &~6.86~ &~23.03~ &~10.29~\\

\multicolumn{4}{c|}{MSDS-RCNN \cite{li_2018_BMVC}}&~12.51~ &22.19 &~8.57~ &~22.32~ &~10.93~ &~20.10~ &~6.93~ &~23.36~ &~10.01~\\
\hline
\hline
\multirow{6}*{AR-CNN} &  \multicolumn{1}{c}{RFA} & \multicolumn{1}{c}{RoIJ}& \multicolumn{1}{c|}{CAF} & \multicolumn{1}{c|}{~} & \multicolumn{1}{c|}{~} &\multicolumn{1}{c|}{~} & \multicolumn{1}{c|}{~} &\multicolumn{1}{c|}{~} &\multicolumn{1}{c|}{~} & \multicolumn{1}{c|}{~} &\multicolumn{1}{c|}{~} &\multicolumn{1}{c}{~}\\
\cline{2-4}
&&&&~12.94~ &~22.27~ &~8.21~ &~14.76~ &~8.92~ &~19.34~ &~6.85~ &~15.62~ &~6.24~\\
&$\checkmark$ &~ &~&~10.90~ &~12.20~ &~2.92~ &~11.85~ &~2.08~ &~16.87~ &~2.01~ &~11.07~ &~1.96~\\
&$\checkmark$&$\checkmark$&~ &~9.87~ &~11.74~ &~1.29~ &~11.01~ &~1.50~ &~15.92~ &~1.51~ &~10.65~ &~1.12~\\
&$\checkmark$&$\checkmark$ &$\checkmark$ &~8.26~ &~9.41~ &~0.98~ &~9.52~ &~1.09~ &~\textbf{9.53}~ &~\textbf{1.02}~ &~9.46~ &~\textbf{0.90}~\\  %原来：9.53, 10.87, 7.31
&\multicolumn{3}{c|}{$+$~ASC}&~\textbf{8.01}~ &~\textbf{9.17}~ &~\textbf{0.93}~ &~\textbf{8.92}~ &~\textbf{0.95}~ &~9.66~ &~1.05~ &~\textbf{8.99}~ &~\textbf{0.90}~\\

\bottomrule
\end{tabular}
\end{center}
\caption{Detection performances and robustness to position shift on the KAIST dataset. 
$O$ refers to MR$^{T}$ scores at the origin, $\mu$ and $\sigma$ denote the mean and standard deviation of MR$^{T}$. We use the re-implemented model of literature \cite{liu2016multispectral, konig2017fully}, and the provided model of literature \cite{zhang2019cross, li_2018_BMVC}.}
\label{table-measure}
\end{table*}

\textbf{Drone-based Dataset} The drone-based dataset is collected from day to night using a DJI  MATRICE 300 RTK drone with pre-calibrated RGB-T cameras, covering different scenes such as city, suburbs, and countryside. It contains $831$ pairs of all-day images with $6,763$ pairs of dense annotations for 4 object categories: car, truck, bus, cyclist. Though the RGB-T cameras are pre-calibrated, there is still the position shift problem due to different physical properties of cameras and external disturbance in operation.

\subsection{Implementation Details}
We use ImageNet \cite{russakovsky2015imagenet} pre-trained VGG-16 \cite{simonyan2014very} as the backbone network.
By default, the $\sigma_{0}$ and $\sigma_{1}$ in RoI jitter is set to $0.05$, which can be adjusted to tackle a wider or narrower position shift. We horizontally flip all the images for data augmentation.
We set the initial learning rate to $0.005$ for 2 epochs and $0.0005$ for 1 more epoch. 
For the drone-based dataset, we use an initial learning rate $0.0005$ for 9 epochs and decay it by 0.1 for another 3 epochs.

We utilize the log-average miss rate (MR) to measure the detection performance, lower score indicates better performance. We plot the miss rate averaged against the false positives per image (FPPI) over the log-space range of $[10^{-2},10^{0}]$. For the test set, we use the widely adopted improved annotation, which is proposed by Liu \etal \cite{liu2016multispectral}. Besides, based on our KAIST-Paired annotation, we use the MR$^{C}$ and MR$^{T}$ to denote the MR for color and thermal modality, respectively. 
For the drone-based dataset, we use mean Average Precision (mAP) to evaluate the performance of detectors.

\subsection{Comparison Experiments}

\textbf{KAIST} The proposed AR-CNN is evaluated and compared to other available competitors \cite{hwang2015multispectral, liu2016multispectral, konig2017fully, li2019illumination, guan2019fusion, zhang2019cross}. As shown in Figure \ref{figure-MR}, our method achieves $9.79$, $8.07$, and $9.03$ MR on the reasonable day, night, and all-day subset respectively, achieving the state-of-the-art performance. Besides, in consideration of the position shift problem, the methods are also evaluated with the KAIST-Paired annotation, \ie MR$^{C}$ and MR$^{T}$. As shown in Figure \ref{figure-MR}, the proposed method has significant advantages, \ie $8.52$ \vs $11.28$ MR$^{C}$ and $8.01$ \vs $12.51$ MR$^{T}$, demonstrating the superiority of our method.

\textbf{CVC-14} Our experiments on the CVC-14 dataset are following the protocol in \cite{Park2018Unified}. As shown in Table \ref{table-cvc}, compared to other competitors, the proposed method achieves the best performance. The greater advantage on the night subset ($18.1$ \vs $30.8$ MR) validates the contribution of thermal modality, and shows that the detection accuracy can be significantly improved by appropriately handling the position shift problem.

\textbf{Drone-based dataset} As shown in Table \ref{table-drone}, compared to the available competitors, our approach achieves the best $36.4$ mAP, which demonstrates the generalization ability from pedestrian detection to the wider object detection.

\begin{table*}[!htbp]
\begin{center}
\linespread{1.2}\selectfont
\begin{tabular}{cccc|c|c|c|c|c|c|c|c|c}
\toprule  
\multicolumn{4}{c|}{\multirow{2}*{Method}} &  \multicolumn{3}{c|}{$S^{0^{\circ}}$} & \multicolumn{2}{c|}{$S^{45^{\circ}}$} & \multicolumn{2}{c|}{$S^{90^{\circ}}$} & \multicolumn{2}{c}{$S^{135^{\circ}}$}\\
\cline{5-13}
&&&&$O$ &$\mu$ &$\sigma$ &$\mu$ &$\sigma$  &$\mu$ &$\sigma$ &$\mu$ &$\sigma$\\

\hline

\multicolumn{4}{c|}{Halfway Fusion \cite{liu2016multispectral}}&~25.10~ &~34.20~ &~7.98~ &~34.80~ &~7.87~ &~32.79~ &~7.36~ &~33.85~ &~8.16~\\

\multicolumn{4}{c|}{Fusion RPN \cite{konig2017fully}}&~20.52~ &~30.22~ &~9.07~ &~29.15~ &~9.23~ &~25.12~ &~6.95~ &~29.97~ &~9.80~\\

\multicolumn{4}{c|}{Adapted Halfway Fusion}&~15.06~ &~ 24.39 ~ &~8.19~ &~25.91~ &~11.32~ &~21.34~ &~7.25~ &~26.58~ &~11.52~\\

\multicolumn{4}{c|}{CIAN \cite{zhang2019cross}}&~14.64~ &~23.86~ &~8.74~ &~24.87~ &~11.08~ &~19.16~ &~6.22~ &~25.20~ &~10.46~\\

\multicolumn{4}{c|}{MSDS-RCNN \cite{li_2018_BMVC}}&~11.28~ &20.06 &~7.95~ &~20.81~ &~9.54~ &~16.61~ &~6.36~ &~21.73~ &~10.14~\\
\hline
\multicolumn{4}{c|}{Ours \cite{Zhang_2019_ICCV}}&~8.86~ &12.51 &~1.60~ &~11.52~ &~2.07~ &~\textbf{11.27}~ &~\textbf{1.41}~ &~11.83~ &~2.22~\\
\multicolumn{4}{c|}{$+$~ASC}&~\textbf{8.52}~ &\textbf{11.79} &~\textbf{1.28}~ &~\textbf{10.58}~ &~\textbf{1.90}~ &~11.31~ &~1.47~ &~\textbf{11.03}~ &~\textbf{2.16}~\\
\bottomrule

\end{tabular}
\end{center}
\caption{The detection robustness to \textit{thermal} position shift (\ie we fix the color image while shifting the thermal image) on the KAIST dataset. MR$^{C}$ is used to evaluate the detection performance.}
\label{table-measure-c}
\end{table*}

\subsection{Robustness to Position Shift}
Most multi-modal systems suffer from the position shift problem, but degrees of the shift are varying with different devices and settings. To clearly set the experiments, we introduce the empirical upper bound of position shift for the weak aligned image pair.

Since the pedestrian detection and object detection tasks use different evaluation metric, \ie log average miss rate (MR, lower is better) and mean average precision (mAP, higher is better), we respectively define the performance degradation rate $\mathrm{R}_{d}$ as follow:
\begin{equation}
\begin{aligned}
\mathrm{R}_{d} = (\mathrm{MR}_{degraded} - \mathrm{MR}_{original}) / \mathrm{MR}_{original} 
\end{aligned}
\end{equation}
\begin{equation}
\begin{aligned}
\mathrm{R}_{d} = (\mathrm{mAP}_{original} - \mathrm{mAP}_{degraded}) / \mathrm{mAP}_{original}
\end{aligned}
\end{equation}

\noindent In this paper, we set the shift of $50\%$ performance degradation ($\mathrm{R}_{d} = 0.5$ ) as the weak aligned bound. For instance, if shifting $10$ and $-15$ pixels along the x-axis leads to $50\%$ performance degradation, then the two upper bounds $\mathrm{B}_{u1}$ and $\mathrm{B}_{u2}$ on this direction are $10$ and $-15$. In experiments, we calculate this bound based on the Halfway Fusion \cite{liu2016multispectral} model.

First, the thermal modality is set as the reference since it usually provides consistent images throughout the day. We use MR$^{T}$ to evaluate the detector. Following the settings in Section \ref{SEC3.3}, the visual results in a surface plot are depicted in Figure \ref{figure-shifting-2}.
As illustrated in Figure \ref{figure-shifting}, compared to the baseline method, the proposed detector significantly enhances the robustness to position shift, and the performance at the origin is also improved.
Besides, we design four metrics to quantitatively evaluate the robustness, \ie $S^{0^{\circ}}$, $S^{45^{\circ}}$, $S^{90^{\circ}}$, $S^{135^{\circ}}$, where the $0^{\circ}$-$135^{\circ}$ indicate the shift directions on the image plane. 
We have $10$ shift modes for each shift direction, which are selected between the origin and weakly aligned bound via equally spacing and rounding. 
Table \ref{table-measure} shows the mean ($\mu$) and standard deviation ($\sigma$) of those $10$ results, our method achieves the best $\mu$ and smallest $\sigma$ on four metrics, demonstrating the robustness to diverse position shift conditions.

\begin{figure*}[!t]
\centering
\includegraphics[width=5.3in]{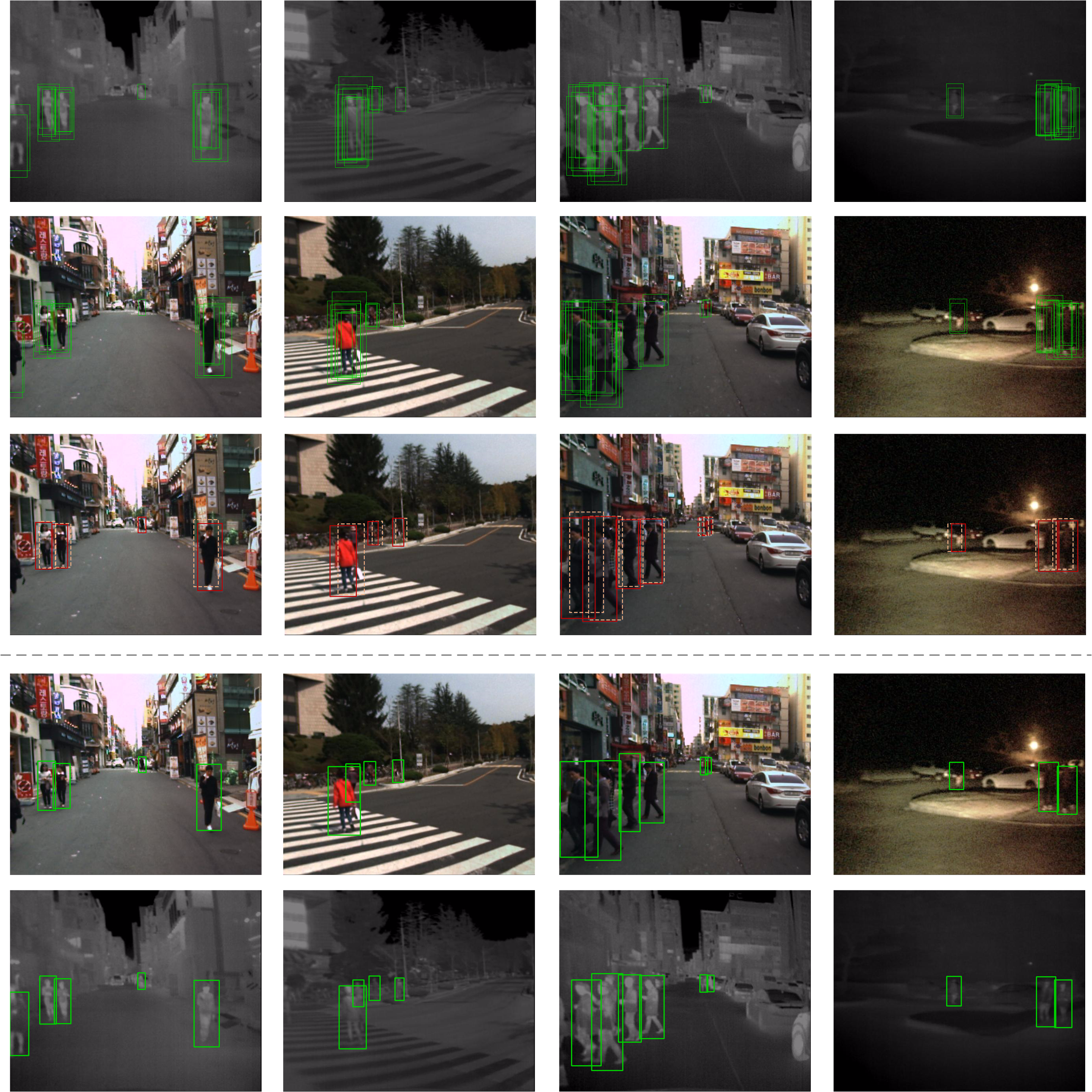}
% where an .eps filename suffix will be assumed under latex,
% and a .pdf suffix will be assumed for pdflatex; or what has been declared
% via \DeclareGraphicsExtensions.
\caption{Qualitative results of the proposed approach. 
The first row illustrates the reference proposals whose confidence score (in the range $[0, 1.0]$) is greater than $0.6$, and the second row shows the corresponding sensed proposals. 
To demonstrate the effectiveness of the Region Feature Alignment module, some proposal instances are shown in the third row: orange dotted boxes denote the reference proposals, which are good ones in the reference image but suffer the position shift problem in the sensed modality; red bounding boxes refer to the adjusted sensed proposals after the region feature alignment process.
In the last two rows, green bounding boxes show the final detection results whose confidence score is greater than $0.6$.}

\label{figure-vis}
\end{figure*}

\textbf{Experiments on the color reference.} Then, we fix the color image as the reference modality. 
Table \ref{table-measure-c} shows that the proposed method still achieves the best performance and smallest standard deviation, further validating the effectiveness of the AR-CNN. 
Additionally, compared to the thermal reference, the color reference configuration performs at a lower level in experiments. 
This validates our intuition: the modality with stable imagery is more suitable to serve as the reference one.

%%% TABLE-hyper-parameter
\begin{table}[!t]
\begin{center}
\linespread{1.15}\selectfont
\begin{tabular}{c|c|c|c}
\toprule  
\multirow{1}*{$\sigma_0$ and $\sigma_1$}& Day & Night & All-day\\

\hline
0.01&~9.97~&~8.13~&~9.29~\\

0.02&~9.90~&\textbf{~8.01}~&~9.10~\\

0.05&~\textbf{9.79}~&~8.07~&~\textbf{9.03}~\\

0.1&~9.96~&~8.17~&~9.21~\\

0.2&~10.07~&~8.22~&~9.36~\\
\bottomrule
\end{tabular}
\end{center}
\caption{Object detection results on different hyper-parameter $\sigma_0$ and $\sigma_1$.}
\label{table-hyper-parameter}
\end{table}

\subsection{Ablation Study}
To analyse our model in more detail, we conduct ablation studies on the KAIST dataset.

\begin{figure*}[!t]
\centering
\includegraphics[width=5.25in]{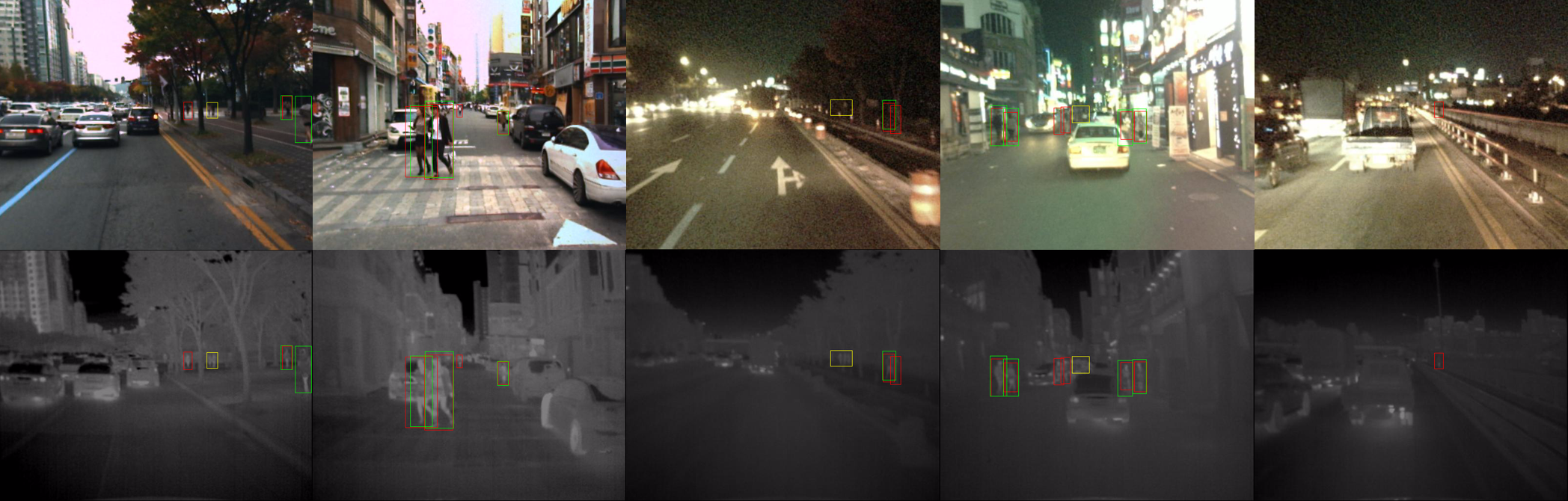}
% where an .eps filename suffix will be assumed under latex,
% and a .pdf suffix will be assumed for pdflatex; or what has been declared
% via \DeclareGraphicsExtensions.
\caption{Visualization of the failed cases on the KAIST test set. Red boxes represent the ground truth, yellow boxes denote ignored objects and green boxes refer to detected objects whose confidence scores are greater than $0.3$. }
\label{figure-failed-cases}
\end{figure*}

\textbf{Region Feature Alignment}
Table \ref{table-measure} shows detection results with and without the RFA module. It can be observed that the miss rate and the $\sigma$ under various position shift conditions are remarkably reduced by the RFA module. Specifically, for $S^{45^{\circ}}$, the RFA reduces the $\sigma$ by a significant $6.84$ (\ie from $8.92$ to $2.08$). For other three metrics, consistent reductions are also observed. Moreover, some visualizations of proposals and detection results are shown in Figure \ref{figure-vis}.
For pedestrians with the position shift problem, proposals of the sensed (color) modality are adjusted to an aligned position. This phenomenon demonstrates that the RFA module can predict the region-wise position shift of two modalities and adaptively adjust the sensed proposals, thus enabling the modality-aligned feature fusion process for better classification and localization.

\textbf{RoI Jitter Strategy} Then we demonstrate the contribution of the RoI jitter. Table \ref{table-measure} shows that the strategy further improves the detector's robustness. Specifically, for $S^{0^{\circ}}$, the $\mu$ is reduced from $12.20$ to $11.74$, and $\sigma$ is decreased from $2.92$ to $1.29$. Meanwhile, it can be observed that this strategy is more helpful for the $\sigma$ than the miss rate at the origin, which demonstrates that the main contribution of this strategy is boosting the robustness to the position shift. Table \ref{table-hyper-parameter} further shows the detection results with varying hyper-parameter $\sigma_0$ and $\sigma_1$. When $\sigma_0$ and $\sigma_1$ are set to 0.05, we can obtain the best overall score. When varying the hyper-parameter, the proposed method exhibits stable performances which are better than the baseline. 

\textbf{Confidence-Aware Fusion}
We also compare performances with and without the CAF module to validate the effectiveness of our multi-modal fusion scheme. Table \ref{table-measure} shows that this module significantly reduces MR$^{T}$ at the origin by $1.61$, yet slightly suppresses the $\sigma$. This illustrates that the main contribution of the CAF module is improving the detection performance, since it conducts adaptive fusion by paying more attention to reliable features. 

Besides, we also visualize some failed cases in Figure \ref{figure-failed-cases} to bring insights for the error analysis and further improvement of the detector. 
Generally, the far-scale tiny people are easier to be missed since it is difficult to obtain sufficient features after the CNN pooling. We also notice that the detection performance will degrade if the environment is complex, e.g. with a cluttered background and a cluster of objects, due to disturbed or incomplete features. In the future, we would like to work on the occluded person by enhancing the classification and localization heads to further improve the detection performance.

%------------------------------------------------------------------------

\begin{figure*}
\centering
\includegraphics[width=5.3in]{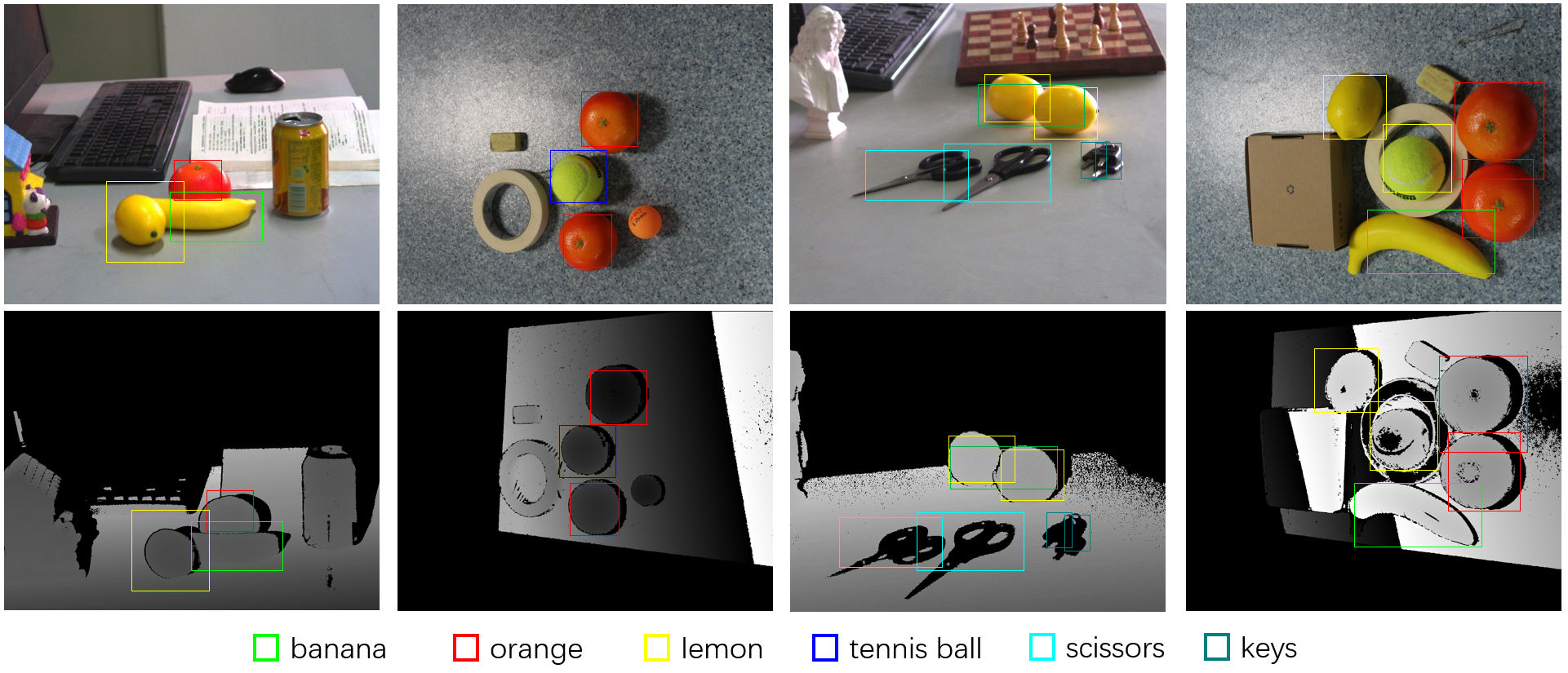}  

\caption{Data instances and detection results in the SL-RGBD dataset. The first row shows RGB images and the second row depicts corresponding depth images that suffer from different degrees of position shift problem.}%
\label{figure-slrgbd}                     
\end{figure*}

\begin{table*}[!htbp]
\begin{center}
\linespread{1.1}\selectfont
\begin{tabular}{cccc|c|c|c|c|c|c|c|c|c}
\toprule  
\multicolumn{4}{c|}{\multirow{2}*{Method}} &  \multicolumn{3}{c|}{$S^{0^{\circ}}$} & \multicolumn{2}{c|}{$S^{45^{\circ}}$} & \multicolumn{2}{c|}{$S^{90^{\circ}}$} & \multicolumn{2}{c}{$S^{135^{\circ}}$}\\
\cline{5-13}
&&&&$O$ &$\mu$ &$\sigma$ &$\mu$ &$\sigma$  &$\mu$ &$\sigma$ &$\mu$ &$\sigma$\\

\hline

\multicolumn{4}{c|}{Amodal3Det (RGB only) \cite{deng2017amodal}}&~30.0~ &~-~ &~-~ &~-~ &~-~ &~-~ &~-~ &~-~ &~-~\\

\multicolumn{4}{c|}{ DSS \cite{song2016deep}}&~36.3~ &~-~ &~-~ &~-~ &~-~ &~-~ &~-~ &~-~ &~-~\\

\multicolumn{4}{c|}{ 3D-SSD \cite{luo20173d}}&~39.7~ &~-~ &~-~ &~-~ &~-~ &~-~ &~-~ &~-~ &~-~\\

\multicolumn{4}{c|}{ Rahman \etal \cite{rahman20193d}}&~\textbf{43.1}~ &~30.2~ &~7.3~ &~31.3~ &~7.1~ &~31.5~ &~6.9~ &~31.0~ &~7.1~\\

\multicolumn{4}{c|}{Amodal3Det \cite{deng2017amodal}}&~40.9~ &~29.2~ &~7.0~ &~30.7~ &~7.0~ &~30.1~ &~6.9~ &~28.7~ &~7.5~\\

\multicolumn{4}{c|}{Halfway Fusion \cite{liu2016multispectral}}&~41.0~ &~29.0~ &~6.7~ &~30.1~ &~7.1~ &~29.8~ &~7.0~ &~29.8~ &~7.2~\\

\multicolumn{4}{c|}{AR-CNN (ours)}&~41.2~ &~\textbf{33.5}~ &~\textbf{3.8}~ &~\textbf{34.3}~ &~\textbf{4.0}~ &~\textbf{34.2}~ &~\textbf{3.9}~ &~\textbf{34.6}~ &~\textbf{4.0}~\\

\bottomrule
\end{tabular}
\end{center}
\caption{Detection performances and robustness to position shift on the NYUv2 dataset. We use mAP for the evaluation. We re-implemented the Halfway Fusion \cite{liu2016multispectral} and Amodal3Det \cite{deng2017amodal} models for comparison.}
\label{table-measure1}
\end{table*}

%------------------------------------------------------------------------
\section{Experiments on Case 2: RGB-D Object Detection}
\label{S6}
In this section, we conduct experiments of 2D object detection on the SL-RGBD dataset and 3D object detection on the NYUv2 \cite{silberman2012indoor} dataset.

\subsection{Dataset}
\textbf{NYUv2} The NYU Depth V2 dataset \cite{silberman2012indoor} is a popular yet challenging dataset for indoor scene understanding, which contains 1,449 RGB-D pairs with 19 indoor object classes. To achieve 3D object detection, \cite{song2015sun} extended the NYUv2 dataset with extra 3D bounding boxes, and refine the depth map by integrating multiple frames of raw video data. In \cite{deng2017amodal}, the 3D annotations are further refined by addressing issues about inconsistency and inaccuracy.
 
\textbf{SL-RGBD} The SL-RGBD dataset is built for indoor recognition, which contains $214$ RGB-D image pairs with $1,297$ object annotations, examples are shown in Figure \ref{figure-slrgbd}. The object classes are selected from a subset of the YCB benchmark, consisting of lemon, banana, strawberries, orange, scissors, plastic bolt and nut, keys. Various patterns of position shift are included to show the characteristics of weakly aligned RGB-D pairs. The details of the collection system are in Section \ref{S3.1}.

\subsection{Implementation Details}
The hyperparameter is set the same as in RGB-T pedestrian detection. We use the ImageNet pre-trained model and fine-tune it on the NYUv2 and SL-RGBD datasets, respectively. We use the same training schedule like that in the all-day RGB-T pedestrian detection.

We use mAP (mean Average Precision) to evaluate the performance of detectors. For 2D and 3D object detection, the detected box is treated as true positive if the IoU is greater than 0.5 and 0.25, respectively. Following previous works \cite{deng2017amodal, luo20173d, rahman20193d}, we use the improved 3D annotation provided by \cite{deng2017amodal} for training and testing.

\subsection{Performance and Robutstness}
\textbf{2D Object Detection} We conduct the 2D object detection experiments on our SL-RGBD dataset. As shown in Table \ref{table-slrgbd}, incorporating the depth modality significantly improves the detection performance (58.7 \vs 52.9 mAP), and our AR-CNN model achieves the best 66.3 mAP results since the position shift problem is taken into consideration.

%%% TABLE-SL-RGBD
\begin{table}[h]
\begin{center}
\linespread{1.15}\selectfont
\begin{tabular}{c|c|c}
\toprule  
\multirow{1}*{Input}&\multirow{1}*{Method} & mAP ($\%$) \\

\hline
\multirow{2}*{RGB}&\multirow{1}*{SSD \cite{liu2016ssd}}&~51.6~\\

&\multirow{1}*{Faster R-CNN \cite{ren2015faster}}&~52.9~\\

\hline

\multirow{3}*{RGB+D}&\multirow{1}*{Halfway Fusion \cite{liu2016multispectral}}&~58.7~\\

&\multirow{1}*{CIAN \cite{zhang2019cross}}&~59.3~\\

&\multirow{1}*{AR-CNN (Ours)}&~\textbf{66.3}~\\
\bottomrule
\end{tabular}
\end{center}
\caption{Object detection results on the SL-RGBD dataset. We use mAP to evaluate the performance of detectors.}
\label{table-slrgbd}
\end{table}

\textbf{3D Object Detection} To further validate the effectiveness of our model, we test the performance and robustness of AR-CNN on the RGB-D based 3D object detection task. Similar to that for 2D pedestrians, we alternatively fix the color and depth image, then manually shift the other. Table \ref{table-measure1} shows the results on the standard NYUv2 dataset, which uses color images as the reference modality. Our AR-CNN achieves comparative performance and best robustness (33.5 \vs 30.2 mean performance and 3.8 \vs 6.7 standard variance in $S^{0^{\circ}}$), which demonstrates the validness and generalization ability of the proposed method.

\begin{figure*}[!t]
\centering
\includegraphics[width=5.5in]{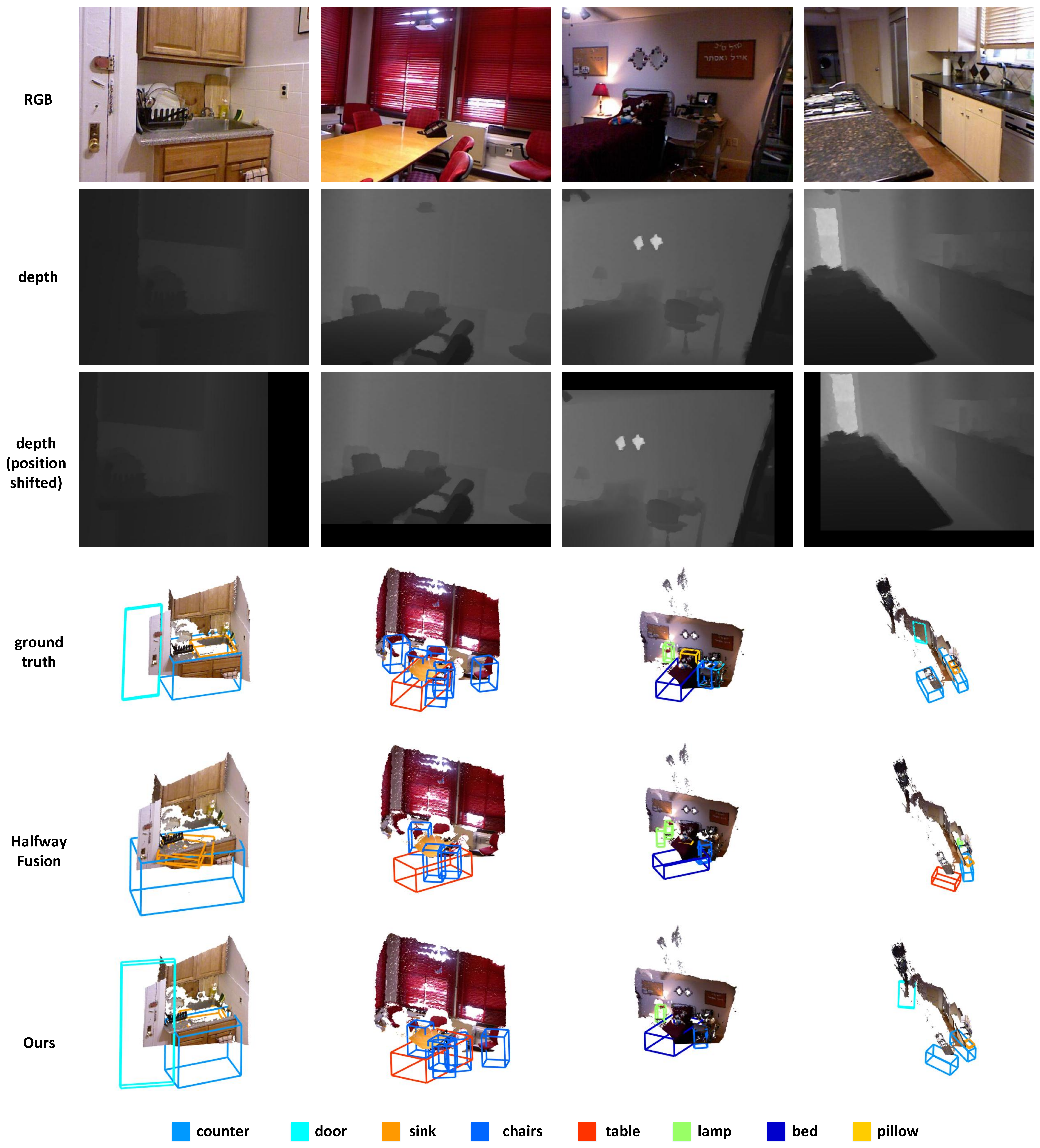}
% where an .eps filename suffix will be assumed under latex,
% and a .pdf suffix will be assumed for pdflatex; or what has been declared
% via \DeclareGraphicsExtensions.
\vspace{0.6em}
\caption{Visualization of results in weakly aligned image pairs. The first and second rows are aligned RGB-D image pairs in which the depth images are refined by integrating multiple frames from the raw video. 
The third row shows different patterns of the position shift problem. The last three rows depict 3D bounding boxes from ground truth, Halfway Fusion, and AR-CNN, respectively.
We show the boxes whose confidence score (in the range $[0, 1.0]$) is greater than $0.6$. For better visualization, the point cloud is generated by aligned RGB and depth images.
}
\label{figure-nyuv2vis}
\end{figure*}

In Figure \ref{figure-nyuv2vis}, we use some examples to further show the model's robustness to position shift problem. When the system suffers from the position shift problem, as shown in the third row, our model achieves better performances than its Halfway Fusion counterpart. For example, the performances of our model and Halfway fusion are 32.1 \vs 23.3 mAP when the shift pattern is $(50,-50)$ on the image plane.

%------------------------------------------------------------------------
\section{Conclusion}
\label{S7}
To handle the practical position shift problem in multi-modal object detection, we propose a novel framework to improve the robustness of the detector when using weakly aligned image pairs.
First, we design a novel region feature alignment module to predict the shift and aligns the multi-modal features. Second, to further enrich the shift patterns, we propose an RoI-level augmentation strategy named RoI jitter. For the all-day RGB-T pedestrian detection, we present new multi-modal labelling named KAIST-Paired and propose the confidence-aware fusion method to pay attention to the more reliable modality. The proposed method achieves the best performance and robustness on CVC-14 and KAIST datasets. For the RGB-D based 2D and 3D detection, we extend our method by aligning the features of 2D proposals and then conducting a more reliable 3D bounding boxes regression. When compared to the state-of-the-art, the proposed method achieves comparative performance and better robustness on NYUv2 and SL-RGBD datasets. 
In the real-world scenario, the weakly aligned characteristic usually exists when using the multi-modal system. The proposed method provides a generic solution for multi-modal object detection, especially when it faces the challenges of the position shift problem.

\ifCLASSOPTIONcaptionsoff
  \newpage
\fi



%\bibliographystyle{IEEEtran}

\bibliographystyle{IEEEtran}
\bibliography{IEEEabrv,mybibfile}

\iffalse

\fi


\vspace{0.5em}
\footnotesize{\textbf{Lu Zhang} received the B.E. degree in automation from Shandong University, in 2016, and the Ph.D. degree from the Institute of Automation, Chinese Academy of Sciences, in 2021, where she is currently an Assistant Professor. Her research interests include computer vision, pattern recognition, and robotics.}

\footnotesize{\textbf{Zhiyong Liu} received the B.E. degree in electronic engineering from Tianjin University, Tianjin, China, in 1997, the M.E. degree in control engineering from the Institute of Automation, Chinese Academy of Sciences (CASIA), Beijing, China, in 2000, and the Ph.D. degree in computer science from the Chinese University of Hong Kong, Hong Kong, in 2003. He is a Professor with the Institute of Automation, CASIA, Beijing. His current research interests include machine learning, pattern recognition, computer vision, and bio-informatics.}

\footnotesize{\textbf{Xiangyu Zhu} received the B.E. degree in Sichuan University (SCU), in 2012, and the Ph.D. degree from the National Laboratory of Pattern Recognition in the Institute of Automation, Chinese Academic of Sciences, in 2017, where he is currently an Associate Professor. His research interests include pattern recognition and computer vision, in particular, 3D morphable model and face analysis.}

\footnotesize{\textbf{Zhan Song} received the Ph.D. degree in mechanical and automation engineering from the Chinese University of Hong Kong in 2008. He is currently with the Shenzhen Institutes of Advanced Technology (SIAT), Chinese Academy of Sciences (CAS), as a Professor. His current research interests include structured light-based sensing, image processing, 3D face recognition, and human-computer interaction.}

\footnotesize{\textbf{Xu Yang} received the B.E. degree in electronic engineering from the Ocean University of China, Qingdao, China, in 2009, and the Ph.D. degree from the Institute of Automation, Chinese Academy of Sciences (CASIA), in 2014, where he is currently an Associate Professor. His current research interests include computer vision, graph algorithms, and robotics.}

\footnotesize{\textbf{Zhen Lei} received the B.S. degree in automation from the University of Science and Technology of China, in 2005, and the Ph.D. degree from the Institute of Automation, Chinese Academy of Sciences, in 2010, where he is currently a Professor. He has published over 100 papers in international journals and conferences. His research interests are in computer vision, pattern recognition, image processing, and face recognition in particular.}

\footnotesize{\textbf{Hong Qiao} received the B.S. and M.S. degrees in engineering from Xi’an Jiaotong University, Xi’an, China, in 1986 and 1989, respectively, the M.Phil. degree from the University of Strathclyde, Glasgow, U.K. in 1997, and the Ph.D. degree in robotics control from De Montfort University, Leicester, U.K., in 1995. She held teaching and research positions with Universities in the U.K. and Hong Kong, from 1990 to 2004. She is currently a “100-Talents Project” Professor with Institute of Automation, Chinese Academy of Sciences, Beijing, China. Her current research interests include robotic manipulation, robotic vision, bio-inspired intelligent robot, and brain-like intelligence.}


%\vfill


\end{document}